\documentclass[10pt,journal,compsoc]{IEEEtran}
\usepackage[hyphens]{url}  
\usepackage{graphicx} 
\usepackage{graphicx}  
\usepackage{amsfonts}
\usepackage{amsmath,amsthm,amssymb}
\usepackage{epsfig}
\usepackage{subfigure}
\usepackage{graphics}
\usepackage{algorithm}
\usepackage{balance}
\usepackage{multirow}
\usepackage{algorithmic}
\usepackage{times}

\usepackage{soul}
\usepackage{url}
\usepackage{bm}
\usepackage{graphicx}
\usepackage{amsmath}
\usepackage{booktabs}
\usepackage{cases}
\usepackage{multicol}
\usepackage{multirow}
\usepackage{makecell}
\usepackage{diagbox}
\usepackage{color}

\newcommand{\tabincell}[2]{\begin{tabular}{@{}#1@{}}#2\end{tabular}}

\newtheorem{theorem}{Theorem}

\newtheorem{Definition}{Definition}
\newtheorem*{Corollary}{Corollary}

\ifCLASSOPTIONcompsoc
   \usepackage[nocompress]{cite}
\else
  \usepackage{cite}
\fi

\ifCLASSINFOpdf
\else
\fi

\begin{document}

\title{Partial Differential Equations Meet \\Deep Neural Networks: A Survey}

\author{Shudong Huang$^*$, 
        Wentao Feng$^*$,~\IEEEmembership{Member, IEEE},
        Chenwei Tang,~\IEEEmembership{Member, IEEE},
        Zhenan He,~\IEEEmembership{Member, IEEE},
        Caiyang Yu,
        and 
        Jiancheng Lv,~\IEEEmembership{Senior Member, IEEE}
\IEEEcompsocitemizethanks{
\IEEEcompsocthanksitem S. Huang, W. Feng, C. Tang, Z. He, C. Yu and J. Lv are with the College of Computer Science, Sichuan University, Chengdu 610065, P. R. China.
\protect\\
E-mail: \{huangsd, Wtfeng2021, tangchenwei, zhenan, lvjiancheng\}@scu. \\ edu.cn,
yucy@stu.scu.edu.cn.
}
\thanks{$^*$equal contribution}
\thanks{Manuscript received Sep. 10, 2022.
This work was partially supported by the State Key Program of the National Natural Science Foundation of China under Grant 61836006, the National Natural Science Foundation of China under Grant 62106164, 62106161, 62076172, the Fundamental Research Funds for the Central Universities under Grant 2022SCU12072, the 111 Project under Grant B21044, the Sichuan Science and Technology Program under Grants 2021ZDZX0011, 2022YFG0188, and the CNPC Innovation Found under Grant 2021DQ02-0903.
}
}


\markboth{Journal of \LaTeX\ Class Files,~Vol.~14, No.~8, August~2015}%
{Shell \MakeLowercase{\textit{et al.}}: Bare Demo of IEEEtran.cls for Computer Society Journals}

\IEEEtitleabstractindextext{%
\begin{abstract}
Many problems in science and engineering can be modeled mathematically by a group of Partial Differential Equations (PDEs). Mechanism-based Scientific Computing following PDEs has long been an essential paradigm for studying topics such as Computational Fluid Dynamics, multiphysics simulation, molecular dynamics, or even dynamical systems. It is a vibrant multi-disciplinary field of increasing importance and with extraordinary potential. At the same time, solving PDEs efficiently has been a long-standing challenge. 
Generally, apart from a few Differential Equations where analytical solutions are directly available, more equations must rely on numerical approaches to be solved approximately, e.g., the Finite Difference Method and Finite Element Method.
These numerical methods usually divide a continuous problem domain into discrete grids and then concentrate on solving the system at each of those points or elements. Although these traditional numerical methods show effectiveness in solving PDEs, the vast number of iterative operations accompanying each step forward greatly reduces the efficiency.
Recently, another equally important paradigm, data-based computation represented by Deep Learning, has emerged as an effective means of solving PDEs. Surprisingly, a comprehensive review of this interesting subdivision of AI for Science is still lacking. 
This survey aims to categorize and review the current progress on Deep Neural Networks for PDEs. We discuss the literature published in this subfield over the past decades and present them in a common taxonomy, followed by an overview and classification of applications of these related methods in scientific research and engineering/medical scenarios. The origin, developing history, character, and sort, as well as the future trends in each potential direction of this area are also introduced.   
\end{abstract}
}
\maketitle
\IEEEdisplaynontitleabstractindextext
\IEEEpeerreviewmaketitle
\IEEEraisesectionheading{\section{Introduction}
\label{sec:introduction}}
\IEEEPARstart {A}{s} a widely-used and rapidly-developed technology, Deep Learning (DL)~\cite{lecun2015deep} powers numerous aspects of modern society, and has dramatically improved the state-of-the-art in data mining, computer vision, natural language processing, etc. DL aims to discover/fit intricate structure in data by training computational models that consists of multiple processing layers to learn the representations of data with multiple levels of abstraction. Beyond the traditional Artificial Intelligence (AI) and Machine Learning (ML) tasks (e.g., speech recognition, object detection, etc.), recent development and tendency of DL has come to explore intersections of multiple disciplines including applied mathematics, Scientific Computing (SC), dynamical systems, etc.~\cite{raissi2020hidden,yadav2015introduction}. 

In the long past development, SC in research works is generally oriented to problems with well-defined mechanisms, such as simulation and prediction based on diverse physical models~\cite{heath2018scientific}. Over time, this kind of method has gradually developed a well-established mode of disposal: by employing instruments such as Differential Equations (DEs) to accomplish descriptive and effective mathematical modeling of mechanisms and applying efficient algorithms to solve them~\cite{Dahlquist2008NMinSC}. In this way, accurate results can be obtained without relying on data or only on a small amount of data. On the contrary, data-based AI computing could acquire approximate solutions without knowing the intrinsic mechanism. Hence, it has become increasingly valuable in this data and computing power explosion era by providing enough data and effective features to train numerous feature-oriented models, such as Artificial Neural Networks (ANNs) in ML~\cite{DOE2020A4S}.




With the rapid development of computing facilities and the continuous optimization of implementations, AI methods, especially Deep Neural Networks (DNNs), have been considered to upgrade SC~\cite{raissi2020hidden,lu202186}. 
At the same time, constructing pre-trained model in a mechanism-based way is expected to reduce considerable workload and boost the learning performance~\cite{Raissi2019Physics}. 
The intersection of DL and SC enriches the forms of modeling and solution of the target problems, and obviously contributes to a mutual reinforcement of these two fields. 
Accordingly, such a trend of fusion is believed to be subordinate to a fresh and very scorching interdisciplinary area, namely AI for Science (AI4Science)\footnote{https://ai4sciencecommunity.github.io}\cite{weinan2021dawning,liu2021physics}. This field has seen a steady stream of innovative works in recent years destined to have a profound impact, such as AlphaFold~\cite{jumper2021highly} and Deep Potential for Molecular Dynamics~\cite{PhysRevLett.120.143001}, and is therefore highly anticipated by the science and engineering communities.

\begin{figure*}[htbp]
\centering
\includegraphics[width=0.88\textwidth]{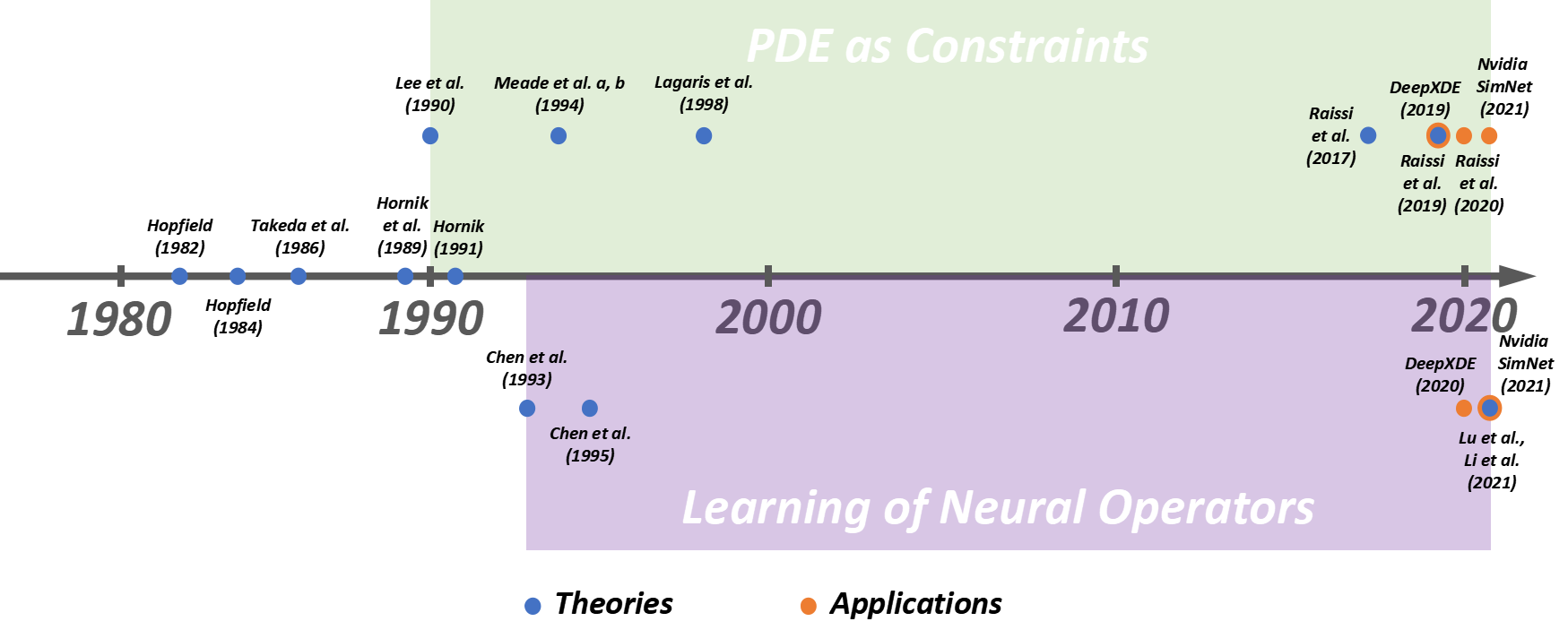}
\caption{Milestones in the development of the theories and applications of the methodologies to solve PDEs using ANNs.}
\label{consGraphs2}
\end{figure*}

\subsection{PDEs and Its Solutions}
In SC, DEs are considered effective for describing various scientific problems and engineering scenarios~\cite{golub1992scientific}. It is comprised of first or higher-order derivatives of unary or multivariate unknown functions, corresponding to the Ordinary Differential Equations (ODEs) or the Partial Differential Equations (PDEs)~\cite{farlow2006introduction}.    
Unlike the algebraic expressions in elementary algebra,  
this approach aims to establish an equation relationship between unknown functions and their derivatives. Its solutions should hence be functions that conform to the relation. 
In practical applications, DEs are essential foundations to depict complex processes in natural and social sciences, such as dynamic systems commonly describing the nexus between states in physical~\cite{Yusupov17} and financial issues~\cite{CHEN20081198} and the stochastic PDEs' being used to recount statistical inferences about random processes~\cite{Rodkina2011}. 




However, solving DEs efficiently has been a long-standing challenge. 
Note that ODEs can be seen as special case of PDEs. 
Only a few simple PDEs are possible to obtain analytical solutions through a combination of a limited number of common operations~\cite{farlow2006introduction}.  
For more complex PDEs, such as the famous Navier-Stokes (N-S) equations, only numerical methods, e.g., the Finite Difference Method (FDM) and the Finite Element Method (FEM), can be applied to acquire the certain-precise solutions relying on high-performance computers. 
At the same time, the multiple iterative computations required by traditional numerical methods undoubtedly limit their practical usage.
So far, engineering applications related to solving PDEs numerically are still mainly concentrated in high-end fields such as aircraft design based on Computational Fluid Dynamics (CFD),  weather forecasting, etc. 

To further improve the efficiency of solving PDEs, it has been tried to strengthen the methods from a mathematical point of view, such as introducing spline functions into FEM to develop the spline FEM~\cite{kolman2017bspline,Qin2019abspline}. Similarly, the central idea of FEM is to discretize the solution domain and then approximate the PDEs' solution numerically. Therefore, introducing a more efficient Fourier Transform (FT) or Laplace Transform to achieve the expression has also proved feasible~\cite{Mugler1988FFT,EIAjou2021AdaptingLT}. Recently, DNNs, which are scorching and considered to possess powerful fitting capabilities for multivariate and high-dimensional functions, have also been re-proposed to solve PDEs lightweight and rapidly.


\subsection{Survey Plan}
To the best of our knowledge, Takeda et al.~\cite{Takeda1986NNforComput} first proposed the application of ANNs to perform computations related to matrix inversion and Fourier transformation in the late 1980s. Subsequently, the paper of Lee et al. in 1990~\cite{lee1990neural} elaborated on their works adopting ANNs to solve PDEs. In general, it was believed that relying on neural algorithms for minimization could build highly parallel algorithms, thereby realizing efficient solving of finite difference equations.  
Later, Chen et al. published several papers~\cite{chen1993approximations,chen1995approximation,chen1995universal} describing the use of neural networks (NNs) to approximate continuous functions and nonlinear operators and apply them to dynamic systems. In 1998, summing up the previous experience, Lagaris et al.~\cite{lagaris1998artificial} further developed an approach using NNs to solve ODEs and PDEs. The works of Chen et al. and Lagaris et al. also identified the two most prominent ideas used for solving PDEs with NNs nowadays. However, like other algorithms based on NNs, early equipment's lack of computing power and the complexity of constructing NNs enormously restricted the practical application of these methods. Until recently, the popularity of GPU accelerated computing platforms and high-performance ML frameworks brought them back to life. 

In this Survey, we aim to review, analyze, and integrate a large number of works over the past nearly forty years to solve PDEs with NN methods, associate with their applications to actual engineering scenes. 
This work first surveys the theoretical approaches to solving PDEs with NNs and their applications into specific practices. The milestones in development are shown in Figure~\ref{consGraphs2}, which highlights the time nodes of emergence of the works about approximating functions and functionals~\cite{meade1994solution,chen1993approximations,Lu2021Learning}. 
The collection of different categories of papers in this review is revealed in Table~\ref{collectedpapers1}.
\begin{table*}[htbp]
\centering
\begin{center}
\caption{Collected references for performing this survey}
\label{collectedpapers1} \scalebox{1.0}{
\begin{tabular}{ccccccccc}
\hline\hline
\rule{0pt}{12pt}
&  & \multicolumn{6}{c}{\textbf{Applications}} & \\ 
\cline{3-8}
\rule{0pt}{12pt}
\multirow{1}{*}{\textbf{Methodologies}} & \multirow{1}{*}{\textbf{Theory}} & \multicolumn{4}{c}{\textbf{Single-Physics}} & \multirow{2}{*}{\tabincell{c}{\textbf{Engineering} \\ \textbf{etc.}}} & \multirow{2}{*}{\tabincell{c}{\textbf{Inverse} \\ \textbf{Problem}}} & \multirow{1}{*}{\textbf{New Perspectives}}\\
\cline{3-6}
\rule{0pt}{12pt}
 &  & \textbf{Fluid} & \textbf{Solid} & \textbf{Heat} & \textbf{Others} &  &  &  \\ 
\cline{1-9}
\rule{0pt}{14pt}
\textbf{General} & \tabincell{c}{[22],[23],[34],\\ {[43]}\textasciitilde[49]} &  &  &  &  &  &  & [175]\textasciitilde[179] \\
\rule{0pt}{18pt}
\tabincell{c}{\textbf{PDEs as} \\ \textbf{Constraints}} & \tabincell{c}{[8],[27],[28],\\ {[31]}\textasciitilde[33],[35],\\ {[36]},[50]\textasciitilde[55]} & \tabincell{c}{[2],[37],\\{[67]}\textasciitilde[80]} & \tabincell{c}{[81],\\{[84]}\textasciitilde[92]} & [82],[93] & \tabincell{c}{[83],\\{[94]}\textasciitilde[98]} & [99]\textasciitilde[131] & \tabincell{c}{[2],[37],[69],\\{[132]}\textasciitilde[148]} & \tabincell{c}{[180]\textasciitilde[186],\\ {[188]}\textasciitilde[191],[196]} \\
\rule{0pt}{14pt}
\tabincell{c}{\textbf{Neural} \\ \textbf{Operator}} & \tabincell{c}{[24]\textasciitilde[26],[29],\\ {[41]},[42],[56]} & [149]\textasciitilde[151] & [152],[153] & [158] & [154]\textasciitilde[157] & [158]\textasciitilde[169] & [170]\textasciitilde[174] & [187],[192]\textasciitilde[195] \\
\rule{0pt}{14pt}
\textbf{Others} & \tabincell{c}{[30],[38]\textasciitilde[40],\\{[57]}} &  &  &  &  &  &  & [197]\textasciitilde[222] \\
\hline\hline
\end{tabular}}
\end{center}
\end{table*}


The reviewed papers were published in the period from 1982 to 2022. Solving PDEs with NNs has renewed attention since the method termed Physics-Informed Neural Networks (PINNs)~\cite{Raissi2019Physics} were proposed in 2019. Thanks to the emergence of high-performance frameworks such as TensorFlow, Pytorch, and PaddlePaddle, PDEs as soft physical constraints to train method for DNNs was quickly applied to the simulation of complex cases. 
Meanwhile, researchers have devoted themselves to approximate multi-precision, multi-scene settings with high-efficiency operators~\cite{Lu2021Learning}, considering that the specifically designed networks based on functional theory is able to depict the families of PDEs. 

The following items were focused on during the literature review: 
\begin{itemize}
\item The theoretical methods proposed in related literature, i.e., how the NNs obtain adequate information from PDEs to help solve them. The types of DNNs for approximating PDEs and the corresponding learning strategies and methods are also introduced in detail.
\item The applications in distinct scenarios including various PDEs to describe these scenarios mathematically, the scale of these scenarios in time and space as well as the computational achievements.
\item Our work on picturing an essential side in this subdivision of AI4Science: how to solve PDEs with DNNs efficiently. 
We propose to achieve a better understanding of the key role of DL in powering the SC, and how this traditional field becomes more refreshing when comes to the confluence of these two research paradigms;  
\item The potential trend directions and new perspectives worthy of attention in the near future, which may contribute to improving the fundamental theories or the forms of engineering application.
\end{itemize}

The rest 
is organized as follows: 
Section~\ref{sect2} summarizes a series of theoretical approaches based on DNNs to solve PDEs; Section~\ref{sect3} elaborates on the mathematical basis of each method and analyses their commonalities and differences; Section~\ref{sect4} introduces the application of each approach in various research and engineering fields according to their characteristics; Section~\ref{sect5} discusses the open challenges, further analyzes the potential trend directions and new perspectives; Finally, we conclude this paper in Section~\ref{sect6}.

\section{NN Works for solving PDEs}\label{sect2}
Traditional numerical methods such as FDM and FEM primarily suffer from the temporal and spatial resolution and discrete form of the grid in continuous study domain. That is, for a complex PDE system, the higher descriptive capability usually requires a more delicate discretization. Besides, the large number of iterative computing at each time step further leads to a heavy load. Hence the balance between resolution and computing load/duration plays an essential role in actual applications. 
In recent decades, the computing capability provided by equipment has maintained a momentum of rapid growth. 
Nevertheless, it is expected to improve efficiency so that overoccupied computing resources can be liberated. Thus it is critical to explore the intersections between AI and SC. Unlike traditional numerical methods, AI-based approaches especially DNNs can be trained to express systems or even families of PDEs, exponentially improving computing efficiency in actual utilization. 
Here we first illustrate several methods of solving PDEs with NNs based on different mathematical foundations.

\subsection{PDEs as Constraints}
This kind of approaches typically employs PDEs as the constraints on training DNNs directly~\cite{lagaris1998artificial,Raissi2019Physics}. 
In many real applications, the constraints are usually brought by generalized prior knowledge, which exhibit the form of DEs and are applied to the model training as a part of loss function~\cite{Lu2021DeepXDE}. 
This methodology was first proposed in the late 1980s and was further developed by Lagaris et al.~\cite{lagaris1998artificial} in the following decade. 
Recently, Raissi et al.~\cite{Raissi2019Physics} proposed the iconic derivative termed PINNs, which released the potential of this methodology in various scenarios~\cite{Zhang2019Physics,Meng2020Physics,raissi2020hidden,Jagtap2020Physics}. There are also DL libraries developed for solving DEs, such as DeepXDE~\cite{Lu2021DeepXDE}. 

Regarding PDEs as training constraints is based on an assumption that the multilayer feedforward neural networks (FFNNs) can be treated as universal approximators for high-dimensional functions~\cite{Hornik1989Multilayer}. 
The commonality of training DNNs and solving DEs is that both of them seek the perfect approximations.  
That is to say, whether it is the FEM that applies basis functions to approximate the proper solution in a unit or the DNNs based on a complex structure to fit high-dimensional functions, their computed/predicted values at each unit center/sampling point should be infinitely close to the actual value. For DNNs, the loss function value, also known as the residual, is desired to be close to zero, and for the numerical approach, it is the computing error of the solution. However, this kind of approaches was also called soft constraints~\cite{lu2021physics} at the beginning since the prior knowledge in the form of PDEs only restricts the correctness of the predicted results but does not guide the optimization.

\begin{figure*}[htbp]
\centering
\includegraphics[width=0.85\textwidth]{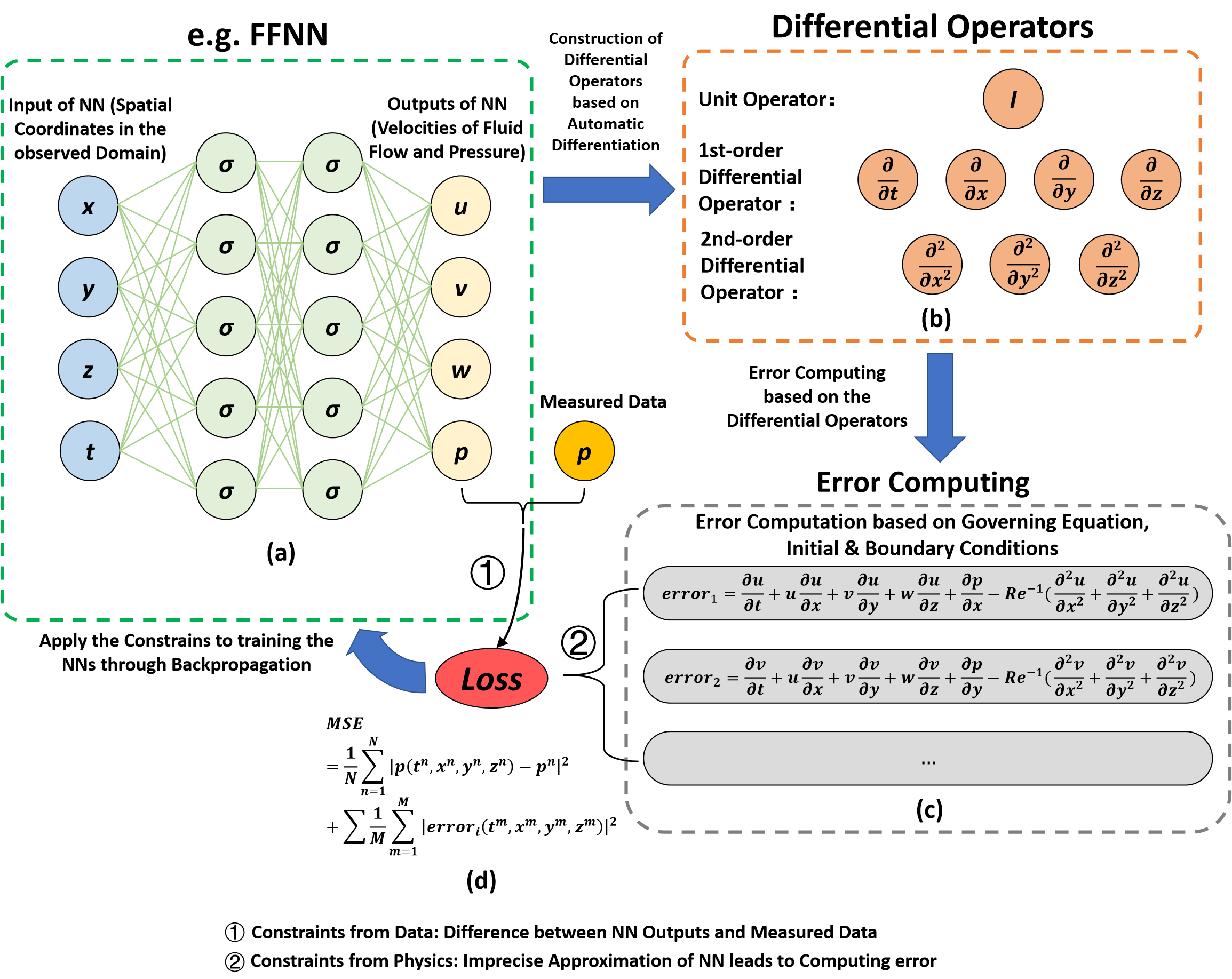}
\caption{Schematic diagram of the PINNs with its application into the solution of Navier-Stokes Equations: (a) the employed feedforward neural networks to accomplish the approximation objectives; (b) and (c) differential operators as components to build up the error computation based on PDEs, this part stands for the constraints from various physical laws; (d) integrated loss function of constraints from physics and data from actual measurement to train the neural networks through backpropagation.}
\label{consGraphs5}
\end{figure*}

Employing PDEs as constraints is generally adopted to accomplish the modeling of some particular scenarios. For example, the Physics-Informed (PI) deep methods have been successfully applied to solve the N-S equations~\cite{raissi2020hidden}, as shown in Figure~\ref{consGraphs5}. Besides, additional restrictions in the form of initial and boundary conditions (ICs/BCs) are taken into account during the modeling. 
Since the constraints based on PDEs are commonly continuous in the study domain, the training of DNNs does not involve the discretized grids as previous numerical methods do~\cite{Cuomo2022ScientificML} and thus can be highly accurate. Moreover, incorporating the posterior information from measured data into restriction or rather loss functions is able to achieve a better solution~\cite{Raissi2018Hidden}. 
However, any modifications in physical parameters, coefficients, or ICs/BCs would lead to different constraints. The shortcomings are hence obvious because of the re-optimization of DNNs for distinct scenarios. 



\subsection{Learning of Neural Operators}
Another kind of approaches is to apply DNNs with unique forms to learn the Neural Operators (NOs). 
With this methodology, DL is able to obtain mesh-free, infinite-dimensional linear and nonlinear operators.  
Motivated by linear algebra~\cite{Banach2009Theory} and functional analysis~\cite{Lax2000FunctionalA,Conway1999ACourse}, Chen et al.~\cite{chen1995approximation,chen1995universal} developed the early prototype of the NO methods based on the operator theory. 
Currently, the two most popular specific instances of this direction are Deep Operator Network (DeepONet), investigated by Lu et al.~\cite{Lu2021Learning}, and Fourier Neural Operator (FNO), proposed by Li et al.~\cite{LiZY2021Fourier}. 
The former developed mathematical and computational instruments from the perspective of building parallel network structures. 
The latter provided its theoretical foundation and a unique design with the help of the FT. 

\begin{figure*}[htbp]
\centering
\includegraphics[width=0.95\textwidth]{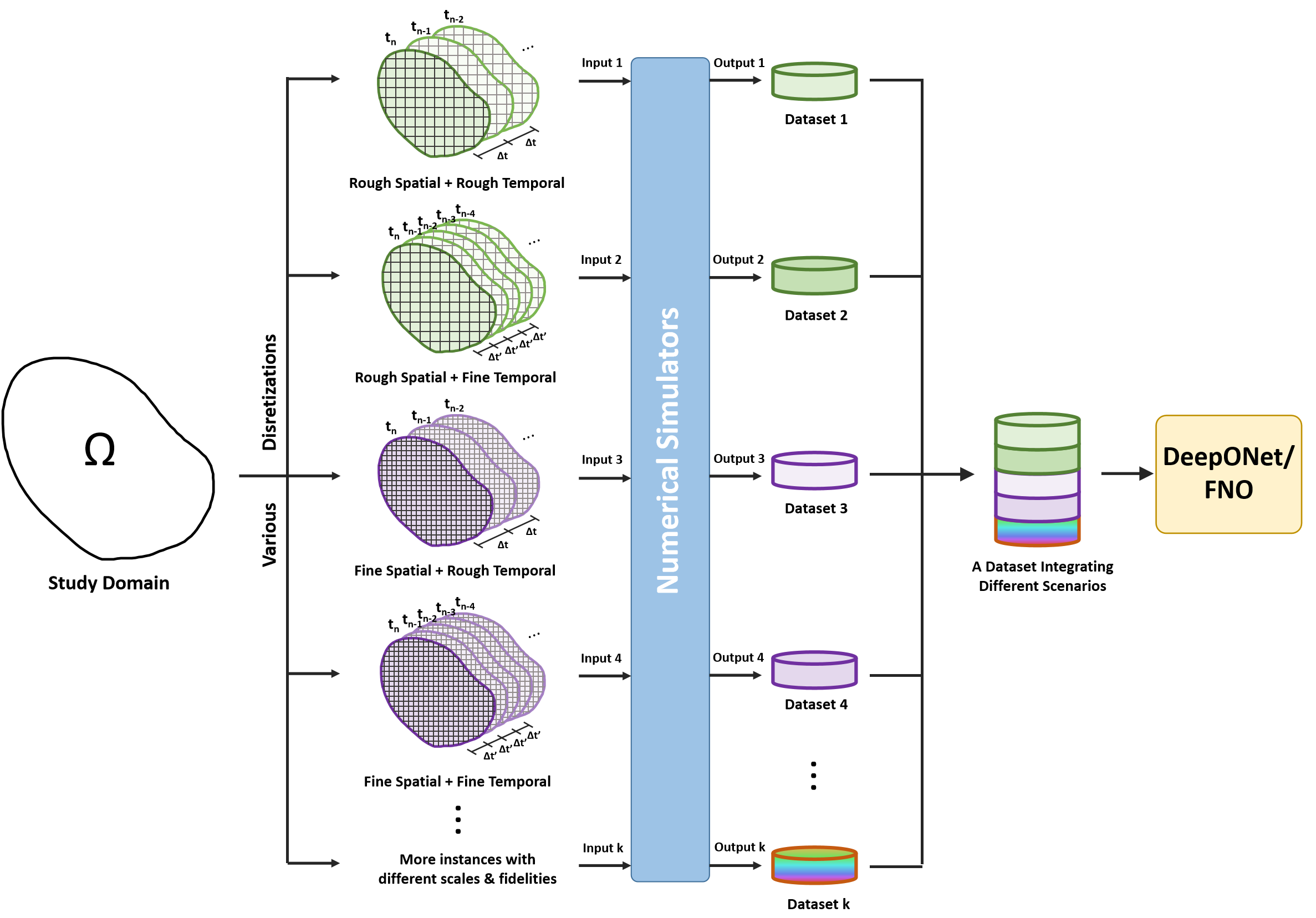}
\caption{Schematic diagram of generating the dataset that can depict different scenarios with various scales and fidelities to train the NO networks.}
\label{consGraphs8}
\end{figure*}

From the theoretical perspective, the NO approaches mainly utilize the parametric DNNs to efficiently express the linear and nonlinear operators that can reveal the commonality in the same PDE family. 
Specifically, DeepONet introduces deep networks and other more complex and domain-specific structures, e.g. Convolutional Neural Networks (CNNs), into the bifurcated network raised by Chen et al. in 1993~\cite{Lu2021Learning}. The feature representation capability of these advanced networks, combined with the thought of approximating functionals by the prototype network, has injected endless vitality into NOs.
As for the FNO, this innovative approach drawing its inspiration from the well-known FT. 
Since FT inherently plays an important role in a variety of applications where approximation or rather fitting is required, FNO has gained excellent characteristics in approximate expression~\cite{LiZY2021Fourier}. 
Kovachki et al.~\cite{Kovachki2021OnUniversal} proved the theoretical guarantees after conducting rigorous mathematical analysis on them.


Based on the approximating of functionals that can depict the PDE families, DNN's parameters learned by handling PDE's multi-scale, multi-fidelity scenarios can effectively eliminate the shackles of the discrete scheme on traditional methods. Besides, the methodology makes the modeling done by neural algorithms no longer only for specific scenarios, and DNNs do not need to be retrained when updating the restrictive conditions.
However, until the second half of 2021, most of the known applications of NOs still focused on directly using operators to learn from the output of numerical solvers. This way, the prior knowledge from PDEs is first transferred into a large-scale dataset numerically, and then fed to the NO networks as input, as shown in Figure~\ref{consGraphs8}. This procedure is time-consuming and labor-intensive, while it seems to be superfluous.


\section{Mathematics}\label{sect3}
In Section~\ref{sect2}, according to the different mathematical foundations on which the approaches are based and the diverse learning objectives, the methodology relying on DNNs to solve PDEs have been classified into two categories, namely the consideration of PDEs as soft constraints and NOs. This section will again review their mathematical principles and go back to the years when primitive conceptions were first proposed (see Table~\ref{collectedpapers2}). It is expected to draw certain enlightenment from the emerging and the current progress on the application of neural algorithms for solving PDEs.

\begin{table}[!htbp]
\begin{center}
\caption{Collected references for reviewing the development history and theoretical background of different methodologies}
\label{collectedpapers2} \scalebox{1.0}{
\begin{tabular}{cc}
\hline\hline
\rule{0pt}{12pt}
\textbf{Methodologies} & \textbf{References} \\ 
\cline{1-2}
\rule{0pt}{16pt}
\textbf{General} & \tabincell{c}{[20],[21],[32],\\{[41]}\textasciitilde[47]}\\
\rule{0pt}{16pt}
\textbf{PDEs as Constraints} & \tabincell{c}{[8],[27],[28],[31]\textasciitilde[33],\\ {[35]},[36],[50]\textasciitilde[55]}\\
\rule{0pt}{16pt}
\textbf{Neural Operator} & \tabincell{c}{[24]\textasciitilde[26],[29],\\{[41]},[42],[56]}\\
\rule{0pt}{14pt}
\textbf{Others} & [30],[38]\textasciitilde[40],[57]\\
\hline\hline
\end{tabular}}
\end{center}
\end{table}

\subsection{From Modified Hopfield Model to Approximation of Continuous Functionals}
In the early 1980s, Hopfield et al.~\cite{Hopfield1982Neuralnet,Hopfield1985Neuralcomput} have realized the programming of a certain class of optimization problems with ANNs and applied the trained NNs to solve one of the NP-complete problems, i.e., the Traveling-Salesman problem, to demonstrate NNs' excellent computational performance. Inspired by this thought, Takeda et al. initiated the idea of employing NNs to solve other computational problems and exhibited related achievements in their works of 1986~\cite{Takeda1986NNforComput}.

\subsubsection{Modified Hopfield Model}
As mentioned before, Takeda et al. proposed to advance the highly parallel programming based on NNs. Meanwhile, they have designed the direct and differential asynchronous transition modes (random delays) to arrange better the oscillation problem of discrete-time Hopfield models, which occasionally arises when the interconnection matrices are non-zero. On this basis, when a particular neuron decides to update its state, there is no need to wait for synchronization with the last neuron. It can take advantage of the information provided by other neurons that have updated their states. The relevant experimental results found that the direct mode can speed up minimizing the energy function, while the differential mode can better reduce oscillations~\cite{Takeda1986NNforComput}. 
However, Takeda et al. also pointed out that since the energy function must be a quadratic function of the neuron state variables, thus it can only deal with the linear problems. 

\subsubsection{Neural Algorithms for Solving DEs}
In 1986 and later in 1989, Koch et al.~\cite{Koch1986Analogneuronal} and Lee et al.~\cite{lee1990neural} have extended the works on minimizing the energy function
in continuous and discrete models using Hopfield model, respectively. 
Since then, the Hopfield networks have been widely applied to minimize the energy functions with arbitrary forms. 

Given a general form of the DE to be disposed of
\begin{equation}
\begin{gathered}
F_{i}\left(x_{l}, u_{j}, d_{m}u_{k}, higher\text{-}order\;derivatives\right)=0,
\label{NADE1}
\end{gathered}
\end{equation}
where $F_{i}$ are $N$ non-singular functions indexed by $i$, with $i$=1 to $N$. $d_{m}u_{k}$, $u_{j}$, and $x_{l}$ are first-order partial derivatives, associated dependent, and independent real variables respectively, with $l$, $m$=1 to $M$ and $j$, $k$=1 to $N$. They are all well defined together with the higher-order derivatives. 

Based on Eq.~\eqref{NADE1}, by discretizing the finite domain in M-dimensional space with a unit volume of $h^M$ and applying FDM to solve DEs, Lee et al. gave a general form of the continuous neural algorithm to solve DE 
\begin{equation}
\begin{gathered}
dW_{i{\bf{a}}s}/dt=-A\sum_{k}\bigg\{\partial\left(F_{k\bf{a}}^2\right)/\partial U_{i\bf{a}}
\\
-\sum_{m}D_{m}[\partial\left(F_{k\bf{a}}^2\right)/\partial\left(D_{m}U_{i\bf{a}}\right)]
\\
+\sum_{ml}D_{ml}[\partial\left(F_{k\bf{a}}^2\right)/\partial\left(D_{ml}U_{i\bf{a}}\right)]
\\
+higher\text{-}order\;terms\bigg\}\left(\partial U_{i\bf{a}}/\partial V_{i{\bf{a}}s}\right),
\\
V_{i{\bf{a}}m}=G\left(W_{i{\bf{a}}m}\right).
\label{NADE2}
\end{gathered}
\end{equation}
where $W_{i{\bf{a}}s}$ are intermediate variables, with $\bf{a}$ being a discrete index vector composed of components $a\left(l\right)$, $l$=1 to $M$. Each component $a\left(l\right)$ represents the discrete coordinate for the individual variabele $x_{l}$. $t$ is the time. $A$ is a positive constant. $F_{k\bf{a}}$ are the finite difference equation corresponding to the differential equation~\eqref{NADE1}, $D_m$, $U_{i\bf{a}}$, $D_mU_{i\bf{a}}$, and $D_mlU_{i\bf{a}}$ are discrete quantities corresponding to the continues quantities in Eq.~\eqref{NADE1}, $V_{i{\bf{a}}m}$ are continuous variables bounded between 0 and 1, and $G$ is a monotonically increasing threshold function bounded between 0 and 1.



This is the first time to propose employing neural algorithms for dealing with DE issues. 
Subsequently, Lee et al.~\cite{lee1990neural} have further elaborated that the ICs can be regarded as BCs in the time domain and exhibited an optical way of implementing the newly developed algorithms. 



\subsubsection{Multilayer FFNNs as Universal Approximators}\label{UniversaAppro313}
Also in the late 1980s, Hornik et al. exhibited a series of further proofs for the ANNs' universal approximation theorem~\cite{Hornik1989Multilayer,Hornik1991Approximationcapab}. 
In those works, the multi-neurons and multilayer architecture of FFNNs, rather than the choices of activation functions, were demonstrated to be the decisive factors that endow NNs with the capability of being universal approximators. According to~\cite{Hornik1991Approximationcapab}, the following theorems and Corollaries can be derived
\begin{theorem}\label{theo1}
For every squashing function $\Psi$, every $r$, and every probability measure $\mu$ on $\left({\bf{R}}^{r}, {\bf{B}}^{r}\right)$, $\sum^{r}\left(\Psi\right)$ is uniformly dense on compacta in ${\bf{C}}^{r}$ and $\rho_{\mu}$-dense in ${\bf{M}}^{r}$. 
\end{theorem}


Theorem~\ref{theo1} reveals that standard FFNNs with only a single hidden layer can approximate any continuous function uniformly on any compact set and any measurable function arbitrarily well in the $\rho_{\mu}$ metric, regardless of the squashing function $\Psi$ (continuous or not), the dimension of the input space $r$, and the input space environment $\mu$. Thus, $\sum$ networks are universal approximators~\cite{Hornik1989Multilayer}. Furthermore, Theorem~\ref{theo2} with more corollaries have shown that the multi-output multilayer FFNNs are universal approximators of vector-valued functions
\begin{theorem}\label{theo2}
Let $\{x_{1},\dots,x_{n}\}$ be a set of distinct points in ${\bf{R}}^{r}$ and let $g$: ${\bf{R}}^{r}\rightarrow\bf{R}$ be an arbitrary function. If $\Psi$ achieves 0 and 1, then there is a funciton $f\in\sum^{r}\left(\Psi\right)$ with $n$ hidden units such that $f\left(x_{i}\right)=g\left(x_{i}\right)$, $i\in\{1,\dots,n\}$.
\end{theorem}



\begin{Corollary}
When $\Psi$ is a squashing function, Theorem 1 remains valid for multioutput multilayer classes $\sum_{l}^{r,s}\left({\Psi}\right)$ approximating functions in ${\bf{C}}^{r,s}$ and ${\bf{M}}^{r,s}$ with $\rho_{\mu}$ replaced with $\rho_{\mu}^{s}$, $\rho_{\mu}^{s}\left(f,g\right)\equiv\sum_{i=1}^{s}\rho_{\mu}\left(f_{i},g_{i}\right)$ and with $\rho_{p}$ replaced with its appropriate multivariate generalization.
\end{Corollary}



\subsubsection{Approximations of Continuous Functionals by NNs}

Based on Hahn-Banach theorem and the representation theorem of Riesz, Cybenko~\cite{Cybenko1989Approximation} raised to apply the functional analysis to upgrade NN's approximation of continuous or other kinds of functions defined on compact sets in ${\bf{R}}^{n}$.  
Inspired by the works of Funahashi~\cite{Funahashi1989Approximation} and Hornik et al.~\cite{Hornik1991Approximationcapab}, Chen et al.~\cite{Chen1990Approximation} pointed out that the proof of Cybenko is restricted to be existential.
Besides, they gave a more constructive proof in 1990 that a sufficient condition for the theorem to hold is the boundedness of $\sigma(x)$, which will be discussed later. In their work of 1993, 
Chen et al.~\cite{chen1993approximations} stated several theorems with corresponding proofs, which elaborate on the capability of FFNNs (with single hidden layer) to approximate the dynamic systems or rather continuous functionals, as illustrated by Theorem~\ref{theo3}


\begin{Definition}
The dynamic system $G$ can be viewed as a map from ${\bf{X}}_{1}$ to ${\bf{X}}_{2}$, where ${\bf{X}}_{1}$ (or ${\bf{X}}_{2}$) stands for the set of ${\bf{R}}^{q_{1}}$-valued (or ${\bf{R}}^{q_{2}}$-valued) functions defined in ${\bf{R}}^{n}$. That is, $Gu=v\in {\bf{X}}_{2}$, $\forall{u}\in {\bf{X}}_{1}$. Simultaneously, let ${\bf{x}}\in \bf{X}$, and define a "windowing" operator $W$ by 
\begin{equation}
\begin{gathered}
\left(W_{{\bf{\alpha}},a}x\right)\left(\beta\right)=
\begin{cases}
{x}\left(\beta\right)\quad if \beta\in {\bf{\Gamma}}_{{\bf{\alpha}},a}    \\
0\qquad\quad if \beta\notin{{\bf{\Gamma}}_{{\bf{\alpha}},a}} 
\end{cases}
\label{Chen19935}
\end{gathered}
\end{equation}
\end{Definition}
\noindent where ${\bf{\alpha}}=\left(\alpha_{1},\dots,\alpha_{2}\right)\in{\bf{R}}^{n}$ and ${\bf{\Gamma}}_{{\bf{\alpha}},a}=\{{\bf{r}}=\left(r_{1},\dots,r_{n}\right)$ $\in{\bf{R}}^{n}, |r_{j}-\alpha_{j}|\leqslant a, \forall{j}=1,\dots,n\}$, that is, $W_{{\bf{\alpha}},a}$ is a "windowed version" of $x$ with the n-dimensional window centered at $\bf{\alpha}$ and its width $2a$.

\begin{Definition}
A map $G$ from ${\bf{X}}_{1}$ to ${\bf{X}}_{2}$ is said to be approximately finite memory, if for all $\epsilon>0$, there is an $a>0$ such that
\begin{equation}
\begin{gathered}
|\left(Gu\right)_{j}\left({\bf{\alpha}}\right)-\left(GW_{\alpha,a}u\right)_{j}\left({\bf{\alpha}}\right)|<\epsilon,\quad j=1,\dots,q_{2}
\label{Chen19936}
\end{gathered}
\end{equation}
holds for any ${\bf{\alpha}}\in{\bf{R}}^{n}$, $u\in{\bf{U}}$.
\end{Definition}
\noindent where $\bf{U}$ is a nonempty set in ${\bf{X}}_{1}$. ${\bf{U}}_{\alpha,a}$ is defined as ${\bf{U}}_{\alpha,a}=\left\{u|{\bf{\Gamma}}_{\alpha,a},u\in{\bf{U}}\right\}$ with $u|{\bf{\Gamma}}_{\alpha,a}$ being the restriction of $u$ to ${\bf{\Gamma}}_{\alpha,a}$, i.e. $u|{\bf{\Gamma}}_{\alpha,a}=W_{\alpha,a}$.  

\begin{theorem}\label{theo3}
If $\bf{U}$ and G satisfy the following assumptions
\\
(1) if $u\in{\bf{U}}$, then $u|{\bf{r}}_{\alpha,a}\in\bf{U}$ for any ${\bf{\alpha}}\in{\bf{R}}^{n}$ and $a>0$;
\\
(2) for all ${\bf{\alpha}}\in{\bf{R}}^{n}$ and $a>0$, ${\bf{U}}_{\alpha,a}$ is a compact set in ${\bf{C}}_{V}$ $(\Pi_{k=1}^{n}[{\alpha}_{k}-a_{k},{\alpha}_{k}+a_{k}])$ or a compact set in ${\bf{L}}_{V}^{p}(\Pi_{k=1}^{n}$ $[{\alpha}_{k}-a_{k},{\alpha}_{k}+a_{k}])$, where $\bf{V}$ stands for ${\bf{R}}^{q_{1}}$;
\\
(3) let $\left(Gu\right)\left({\bf{\alpha}}\right)=\left(\left(Gu\right)_{1}\left({\bf{\alpha}}\right), \dots, \left(Gu\right)_{q_{2}}\left({\bf{\alpha}}\right)\right)$, then each $\left(Gu\right)_{j}\left({\bf{\alpha}}\right)$ is a continuous functional defined over ${\bf{U}}_{\alpha,a}$, with the corresponding topology in ${\bf{C}}_{V}(\Pi_{k=1}^{n}[{\alpha}_{k}-a_{k},{\alpha}_{k}+a_{k}])$ or ${\bf{L}}_{V}^{p}$ $(\Pi_{k=1}^{n}$ $[{\alpha}_{k}-a_{k},{\alpha}_{k}+a_{k}])$,\vspace{0.3em}
\\
and $G$ is of approximately finite memory, then for any $\epsilon>0$, there exist $a>0$, a positive integer $m$, $\left(m+1\right)^{n}$ points in $\Pi_{k=1}^{n}$ $[{\alpha}_{k}-a_{k},{\alpha}_{k}+a_{k}]$, a positive integer $N$, constants $c_{i}\left(G, {\bf{\alpha}}, a\right)$ depending on $G$, $\bf{\alpha}$, $a$ only, and $q_{2}\times\left(m+1\right)^{n}$-vectors $\bar{\xi}_{i}$, $i=1$, $\dots$, $N$, such that
\begin{equation}
\begin{gathered}
\Bigg|\left(Gu\right)_{j}\left({\bf{\alpha}}\right)-\sum_{i=1}^{N}c_{i}\left(G, {\bf{\alpha}}, a\right)
\\
\sigma\left({\bar{\xi}}_{i}\cdot {\bar{u}}_{q_{1},n,m}+\theta_{i}\right)\Bigg|<\epsilon,\quad j=1,\dots,q_{2}
\label{Chen19937}
\end{gathered}
\end{equation}
or
\begin{equation}
\begin{gathered}
\Bigg|\left(Gu\right)_{j}\left({\bf{\alpha}}\right)-\sum_{i=1}^{N}c_{i}\left(G, {\bf{\alpha}}, a\right)
\\
\sigma\left({\bar{\xi}}_{i}\cdot {\bar{u}}_{q_{1},n,m}^{*}+\theta_{i}\right)\Bigg|<\epsilon,\quad j=1,\dots,q_{2}
\label{Chen19938}
\end{gathered}
\end{equation}
\end{theorem} 
\noindent where ${\bar{u}}_{q,n,m}=\left(u_{l}\left(x_{1}^{j_{1}}, \dots, x_{n}^{j_{n}}\right)\right)$, $l=1$, $\dots$, $q$, $j_{k}=0$, $1$, $\dots$, $m$, $k=1$, $\dots$, $n$ are $q\times\left(m+1\right)^n$-vectors.
${\bar{u}}_{q,n,m}^{*}$ are vectors of the same dimension obtained by replacing $u_{l}\left(x_{1}^{j_{1}}, \dots, x_{n}^{j_{n}}\right)$ in ${\bar{u}}_{q,n,m}$ by ${\left(\frac{1}{2h}\right)}^{n}\idotsint_{x_{k}^{j_{k}}-h\leqslant x_{k}\leqslant x_{k}^{j_{k}}+h}$ $u_{l}\left(x_{1}, \dots, x_{n}\right)dx_{1}\dots dx_{n}$.


The formulation in Theorem~\ref{theo3} has been very close to that of the nonlinear operators arising later. Moreover, Chen et al. finished a series of extended works on the existing foundations in the following two years.

\subsection{ANNs for Solving DEs}
Solving DEs with ANNs has been attracted much attention over the past few decades. 
As mentioned before, one of the earliest work
can be traced back to the proposal of neural algorithm for solving DEs in 1990~\cite{lee1990neural}, in which the first and higher order DE can be discretized using finite difference techniques, and the resulting algebraic equations are transformed into an energy function that minimized by the Hopfield NNs. 

\subsubsection{FFNNs for Solving Linear ODEs}
Instead of solving DEs by dealing with the finite difference equations, Meade and Fernandez~\cite{meade1994numerical,meade1994solution} subsequently proposed to approximate arbitrary linear ODEs with FFNNs, which can be treated as a prototype of  PINNs~\cite{Raissi2019Physics}. In detail, the solution of ODEs is approximated by combining piecewise polynomial splines, where the combination parameters are determined by training a FFNN. Considering that the mathematical model of physical processes can be directly and accurately incorporated into the FFNN architecture, approaches in~\cite{meade1994numerical,meade1994solution} also impose the ICs and BCs on NNs’ parameters.
In~\cite{gobovic1993design}, Gobovic and Zaghloul presented a theory for a locally connected set of neural cells to solve PDEs corresponding to the heat flow problem. In detail, the elliptic PDEs with constant coefficients are formulated as an energy function and can be minimized using VLSI CMOS techniques. The electrical circuit is designed as a cellular neural structure where each cell is implemented as a neuron. 
Afterwards, based on further works in~\cite{gobovic1994analog} and~\cite{yentis1994cmos}, \cite{yentis1996vlsi} presented a locally connected NN for solving a class of PDEs, in which with the usage of parallel processing and the possibility of hardware implementation on neural chips. 

\subsubsection{ANNs for Solving Second-Order Nonliear Equations}
According to~\cite{lagaris1998artificial}, the solution of DEs can be written as the sum of two parts in which the first part ensures ICs and BCs in Dirichlet and/or Neumann forms while the second part is established so as not to affect the conditions. With the help of ANNs, Lagaris et al.~\cite{lagaris1998artificial} proposed a solution of the second-order nonlinear equations up to seven decimal digits precision. As a pioneer work of DNNs for PDEs, this approach would be briefly clarified. 

Consider a general form of DE to be solved 
\begin{equation}\label{MLPNNforPDE1}
F\left(\vec{x}, y(\vec{x}), \nabla y(\vec{x}), \nabla^{2} y(\vec{x})\right)=0, \ \bar{x} \in D
\end{equation}
\noindent where $\vec{x}=\left(x_{1}, x_{2},\ldots, x_{n}\right) \in R^{n}$ subject to certain BCs, $D \subset R^{n}$ represents the definition domain, and $y(\vec{x})$ denotes the solution to be calculated. 

In order to obtain a solution to Eq.~\eqref{MLPNNforPDE1}, the first step is to discretize domain $D$ and the corresponding boundary $S$ into a set of discrete points $\hat{D}$ and $\hat{S}$, respectively. Then DE
can be transformed into a system of equations defined on the constraints imposed by BCs, as illustrated by Eq.~\eqref{MLPNNforPDE2}
\begin{equation}\label{MLPNNforPDE2}
F\left(\vec{x}_{i}, y\left(x_{i}\right), \nabla y\left(x_{i}\right), \nabla^{2} y\left(x_{i}\right)\right)=0 \quad \forall \quad \vec{x}_{i} \in \hat{D}
\end{equation}
\noindent where $y_{t}(\vec{x},\vec{p})$ denotes a trial solution with adjustable parameters $\vec{p}$. 

Thus, the problem can be transformed to Eq.~\eqref{MLPNNforPDE3} subject to the constraints imposed by BCs
\begin{equation}\label{MLPNNforPDE3}
\min _{\vec{p}} \sum_{x_{i} \in \hat{D}} F\left(\left(\vec{x}_{i}\right), y_{t}\left(\vec{x}_{i}, \vec{p}\right), \nabla y_{t}\left(\vec{x}_{i}, \vec{p}\right), \nabla^{2} y_{t}\left(\vec{x}_{i}, \vec{p}\right)\right)^{2}
\end{equation}

Assume that the trial solution $y_{t}(\vec{x})$ satisfies the given ICs/BCs, it can be achieved by a sum of two terms
\begin{equation}\label{MLPNNforPDE4}
y_{t}(\vec{x}) = A(\vec{x}) + f(\vec{x}, N(\vec{x}, \vec{p})) 
\end{equation}
\noindent where the first term $A(\vec{x})$ contains no adjustable parameters, and the second term $N(\vec{x}, \vec{p})$ employs a FFNN whose weights and biases are to be adjusted to handle the minimization problem. 

Up to this point, the study issue has been transformed from the original constrained optimization problem into an unconstrained one, which is obviously easier to solve due to the choice of the form of $y_{t}(\vec{x})$ that satisfies by construction BCs.

\begin{figure*}[htbp]
\centering
\includegraphics[width=0.93\textwidth]{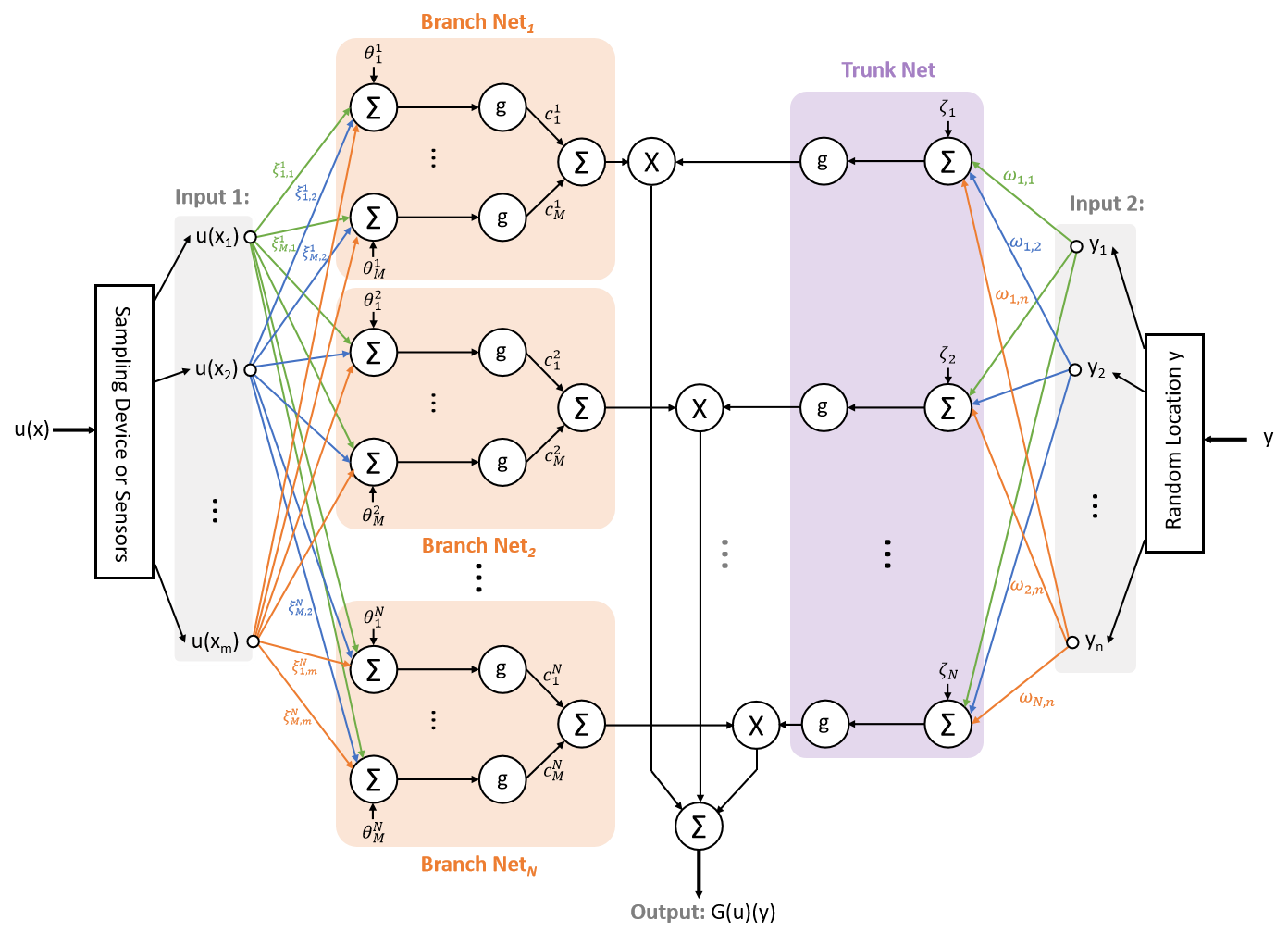}
\caption{The bifurcated architecture of a neural network to approximate nonlinear operator $G\left(u\right)\left(y\right)$ based on Theorem~\ref{theo4}, with marking the Branch Nets and Trunk Net proposed in the stacked configuration of DeepONet.}
\label{consGraphs3}
\end{figure*}

\subsubsection{PINNs}
In recent years, inspired by Lagaris et al.~\cite{lagaris1998artificial} and the idea that considers DNNs as universal function approximators~\cite{Hornik1991Approximationcapab}, Raissi et al.~\cite{Raissi2019Physics} applied the automatic differentiation (AD) technique~\cite{baydin2018automatic} to differentiate the input coordinates and model parameters of NNs, thus establishing the method of PINNs. Specifically, this approach allows numerous physical laws depicted by PDEs to be applied as a part of loss functions for DNNs, thereby effectively regularizing the optimization in the training procedure~\cite{Raissi2018Hidden}. As mentioned before, the training of PINNs does not require discretization of the temporal or spatial domains. At the same time, their practical usage avoid a large number of iterative computations necessary for numerical approaches such as FEM. As one of the hallmarks of the current development, this approach would be illustrated in detail.

First consider the parameterized and nonlinear PDEs possessing the general form
\begin{equation}
u_t+\mathcal{N}[u ; \lambda]=0, x \in \Omega, t \in[0, T]
\end{equation}
\noindent where $u(t, x)$ denotes the hidden solution. $\mathcal{N}[\cdot ; \lambda]$ is a nonlinear operator by $\lambda$. $\Omega$ is a subset of ${\bf{R}}^D$. 

With the setup, a wide range of physical issues can be covered. Further, the problems can be classified into two categories based on whether the parameter $\lambda$ needs to be adjusted according to the observed data.  Considering a realistic scenario of incompressible fluid flow integrating both issues~\cite{Raissi2019Physics}, an explicit form of the well-known N-S equations in two dimensions is given by Eq.~\eqref{MLPNNforPDE5}.
\begin{equation}\label{MLPNNforPDE5}
\begin{aligned}
&u_t+\lambda_1\left(u u_x+v u_y\right)=-p_x+\lambda_2\left(u_{x x}+u_{y y}\right), \\
&v_t+\lambda_1\left(u v_x+v v_y\right)=-p_y+\lambda_2\left(v_{x x}+v_{y y}\right)
\end{aligned}
\end{equation}
\noindent where $u(t, x, y)$ denotes the x-component of the velocity field, $v(t, x, y)$ the y-component, and $p(t, x, y)$ the pressure. $\lambda=\left(\lambda_1, \lambda_2\right)$ are the unknown parameters.

Since the incompressible flow follows the mass conservation and considering that the velocity field possesses noisy measurement $\left\{t^i, x^i, y^i, u^i, v^i\right\}_{i=1}^N$, after defining $f(t, x, y)$ and $g(t, x, y)$ (see Eq.~\eqref{MLPNNforPDE6}), PINNs can be trained by minimizing the mean squared error (MSE) loss given in Eq.~\eqref{MLPNNforPDE7}. 
\begin{equation}\label{MLPNNforPDE6}
\begin{aligned}
&f:=u_t+\lambda_1\left(u u_x+v u_y\right)+p_x-\lambda_2\left(u_{x x}+u_{y y}\right), \\
&g:=v_t+\lambda_1\left(u v_x+v v_y\right)+p_y-\lambda_2\left(v_{x x}+v_{y y}\right)
\end{aligned}
\end{equation}
\begin{equation}\label{MLPNNforPDE7}
\begin{gathered}
MSE:= \\
\frac{1}{N} \sum_{i=1}^N\left(\left|u\left(t^i, x^i, y^i\right)-u^i\right|^2+
\left|v\left(t^i, x^i, y^i\right)-v^i\right|^2\right)\\
+ \frac{1}{N} \sum_{i=1}^N\left(\left|f\left(t^i, x^i, y^i\right)\right|^2+\left|g\left(t^i, x^i, y^i\right)\right|^2\right)
\end{gathered}
\end{equation}
\noindent where $[f(t, x, y)\  g(t, x, y)]$ denotes a PINN.

Based on this innovative approach, Raissi et al. carried out in-depth studies for scenarios containing complex dynamics, such as flow past a circular cylinder or in an intracranial aneurysm, in both their works of~\cite{Raissi2019Physics} and~\cite{raissi2020hidden}.

\subsection{Learning NOs by ANNs}

In 1995, Chen et al.~\cite{chen1995universal} conducted a systematic investigation of NNs' fitting capability. By demonstrating the strengths of NNs to approximate the nonlinear functionals defined on some compact sets of a Banach space and the nonlinear operators, they have proved that this form of computation is able to approximate the output as a whole of a dynamic system.
From this point of view, NOs are undoubtedly closer to the realm of ML than the former approaches and, thus, more potent in terms of performance and usage.

\subsubsection{Universal Approximation to Nonlinear Operators by NNs}
With the operator in Theorem~\ref{theo4}, NNs can accurately depict various characteristics and even numerous details of the dynamic systems. If the network is graphically represented, as Chen et al. have illustrated in their paper~\cite{chen1995universal}, a bifurcated parallel structure can be exploited, as shown in 
Figure~\ref{consGraphs3}. Meanwhile, this model is also a one-layer version of the stacked DeepONet~\cite{Lu2021Learning} (it will be introduced later), which means that the trunk net as well as all the branch nets possess only one hidden layer. 

\begin{theorem}\label{theo4}
Suppose that $g\in{\bf{F}}_{TW}$, $\bf{X}$ is a Banach Space, ${\bf{K}}_{1}\subseteq{\bf{X}}$, ${\bf{K}}_{2}\subseteq{\bf{R}}^{n}$ are two compact sets in $\bf{X}$ and ${\bf{R}}^{n}$, respectively, ${\bf{V}}$ is a compact set in ${\bf{C}}_{{K}_{1}}$, $G$ is a nonlinear continuous operator, which maps $\bf{V}$ into ${\bf{C}}_{{K}_{2}}$, then for any $\epsilon>0$, there are a positive integer $M$, $N$, $m$, constants $c_{i}^{k}$, $\zeta_{k}$, $\xi_{i,j}^{k}\in{\bf{R}}$, points $\omega_{k}\in{\bf{R}}^{n}$, $x_{j}\in{\bf{K}}_{1}$, $i=1$, $\dots$, $M$, $k=1$, $\dots$, $N$, $j=1$, $\dots$, $m$, such that 
\begin{equation}
\begin{gathered}
\Bigg|G\left(x\right)\left(y\right)-\sum_{k=1}^{N}\sum_{i=1}^{M}c_{i}^{k}g\left(\sum_{j=1}^{m}{\xi}_{i,j}^{k}u\left(x_{j}\right)+\theta_{i}^{k}\right)
\\
\cdot g\left({\omega}_{k}\cdot y+{\zeta}_{k}\right)\Bigg|<\epsilon,
\\
\forall{u}\in{\bf{V}}\ \ and\ \ \forall{y}\in{\bf{K}}_{2}
\label{Chen19953}
\end{gathered}
\end{equation}
\end{theorem}

\noindent where ${\bf{F}}_{TW}$ stands for the set of all the Tauber-Wiener functions, ${\bf{C}}_{K}$ is the Banach space of all continuous functions defined on $\bf{K}$, with norm, and $c_{i}\left(f\right)$ is a linear continuous functional defined on $\bf{U}$.

Up to this point, the mathematical foundations for applying NNs to universally approximate nonlinear operators, or the output as a whole of a nonlinear dynamical system, have been laid firmly. However, because of the technical limitations in realizing the massive computing, it took almost another twenty-five years before the relevant approaches began their rapid development.

\subsubsection{DeepONet}
Nowadays, the improvement in computing capability and the emergence of high-performance frameworks have made it more convenient than ever to implement DNN-based ML. As a result, various approaches that had been theoretically proven but were difficult to accomplish at the time are unearthed and realized. With this in mind, Lu et al. integrated DL with Chen et al.'s theorems of approximating nonlinear operators and proposed a more efficient operator learning network termed DeepONet~\cite{Lu2021Learning}. Specifically, considering the vigorous expressibility of deep networks and the unique advantages of different networks in respective fields, such as CNNs for images and the recurrent neural networks (RNNs) for sequential data, they introduced abundant DNNs into the development of DeepONet, to build more diverse and targeted branch and trunk nets (see Theorem~\ref{theo5})
\begin{theorem}\label{theo5}
Suppose that $\bf{X}$ is a Banach Space, ${\bf{K}}_{1}\subseteq{\bf{X}}$, ${\bf{K}}_{2}\subseteq{\bf{R}}^{n}$ are two compact sets in $\bf{X}$ and ${\bf{R}}^{n}$, respectively, ${\bf{V}}$ is a compact set in ${\bf{C}}_{{K}_{1}}$. Assume that $G:{\bf{V}}\rightarrow{\bf{C}}_{{K}_{2}}$ is a nonlinear continuous operator. Then, for any $\epsilon>0$, there are a positive integers $m$, $p$, continuous vector functions ${\bf{g}}:\ {\bf{R}}^{m}\rightarrow{\bf{R}}^{p}$, ${\bf{f}}:\ {\bf{R}}^{n}\rightarrow{\bf{R}}^{p}$, and $x_1$, $x_2$, $\dots$, $x_{m}\in{\bf{K}}_{2}$, such that
\begin{equation}
\begin{gathered}
\Bigg|G\left(x\right)\left(y\right)-\langle\underbrace{{\bf{g}}\left(u\left(x_{1}\right),u\left(x_{2}\right),\dots,u\left(x_{m}\right)\right)}_{branch},
\underbrace{{\bf{f}}\left(y\right)}_{trunk}\rangle\Bigg|<\epsilon,\vspace{0.8em}
\\
\forall{u}\in{\bf{V}}\ \ and\ \ \forall{y}\in{\bf{K}}_{2}
\label{Lu20211}
\end{gathered}
\end{equation}
\end{theorem}
\noindent where $\langle\cdot,\cdot\rangle$ denotes the dot product of branch and trunk net in ${\bf{R}}^{p}$, and the functions $\bf{g}$ and $\bf{f}$ can be chosen as diverse classes of NNs, which satifsy the classical universal approximation theorem of functions, such as fully connected NNs, RNNs and CNNs.

It should be noted that the stacked layout of branch nets was altered to a non-stacked configuration in Theorem~\ref{theo5} so that the computational cost could be reduced. This way, the many branch nets in Chen et al.'s initial work are merged together into a single one~\cite{Lu2021Learning}. In addition, Lu et al. introduced the bias into the branch net and the last stage of operation. The actual results in their paper and the proof by Lanthaler et al.~\cite{Lanthaler2022Errorestimates} have shown that those reformations are beneficial to improving the performance of DeepONet.

\subsubsection{FNO}
At the same time, inspired by Lu et al. and also in conjunction with the study on Graph Kernel Network~\cite{li2020multipole}, Li et al. further brought the FT, which is frequently used in spectral methods to solve DEs, into the realm of NOs~\cite{LiZY2021Fourier}. Up to now, the application of FT has deep roots in the development of DL, including the proof of the universal approximation theorem mentioned in Section~\ref{UniversaAppro313}. Accordingly, Li et al. have proposed an approach of applying the FT-based architecture to learn the nonlinear operators, known as Fourier Neural Operator~\cite{LiZY2021Fourier}. They further parameterized kernel integral operator into Fourier space, and hence proposed the Fourier integral operator, as exhibited by Definition~\ref{defin3}
\begin{Definition}\label{defin3}
Define the Fourier integral operator $K$
\begin{equation}
\begin{gathered}
\left(K\left(\phi\right)v_{t}\right)\left(x\right)=F^{-1}\left({\bf{R}}_{\phi}\cdot\left(Fv_{t}\right)\right)\left(x\right)\quad\forall{x\in{\bf{D}}}
\label{Li20213}
\end{gathered}
\end{equation}
with  
\begin{equation}
\begin{gathered}
\left(Ff\right)_{j}\left(k\right)=\int\limits_Df_{j}\left(x\right)e^{-2i\pi\langle x,k\rangle}dx,
\\
\left(F^{-1}f\right)_{j}\left(x\right)=\int\limits_Df_{j}\left(k\right)e^{2i\pi\langle x,k\rangle}dk,
\label{Li20214}
\end{gathered}
\end{equation}
\end{Definition}

\noindent where $F$ denote the FT of a functiuon $f$: $D\rightarrow{\bf{R}}^{d_v}$ and $F^{-1}$ its inverse, with $j=1$, $\dots$, $d_v$ and $i=\sqrt{-1}$ as the imaginary unit. ${\bf{R}}_{\phi}$ is the FT of a periodic function $\kappa$: $\bar{\bf{D}}\rightarrow{\bf{R}}^{d_{v}\times d_{v}}$, parameterized by $\phi\in\Theta_{K}$.

Apart from this, Li et al. also provided specific elaborations on the implementation of FNO. In the accompanying numerical experiments and also in the proof by Kovachki et al.\cite{Kovachki2021OnUniversal}, the novel NNs reinforced by Fast Fourier Transform (FFT) is proved to be able to learn the highly nonlinear operator more efficiently.


\section{Actual Applications}\label{sect4}
\begin{figure*}[htbp]
\centering
\includegraphics[width=0.87\textwidth]{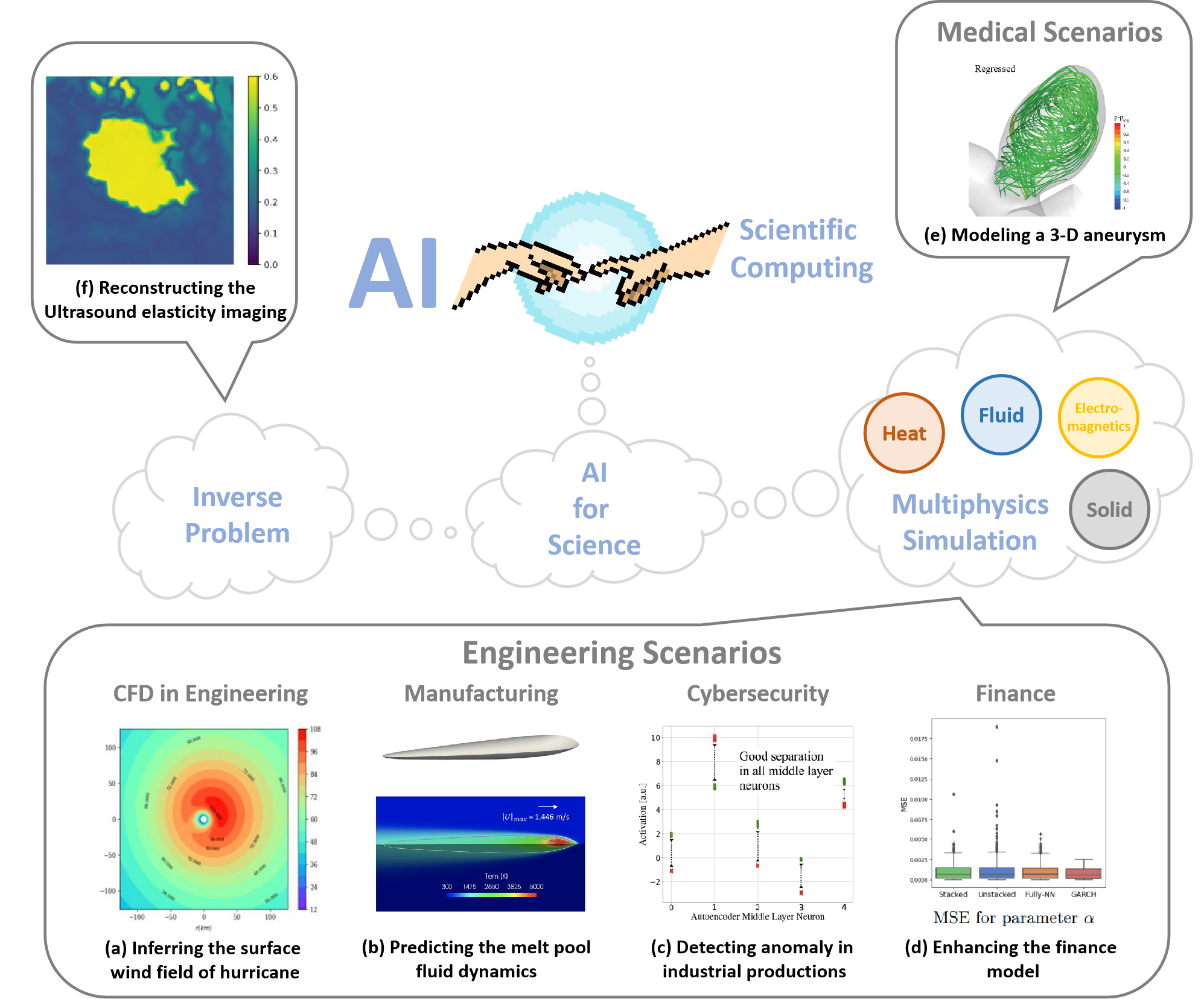}
\caption{Fusion of AI and SC - the emergence of the previously mentioned two kinds of methodologies (employing PDEs as Constraints to train DNNs and constructing NOs) has boosted the development of AI4Science. Nowadays, they are applied in various fields of science and engineering, including the modeling of Single-Physics, Multiphysics, Engineering, and Medical Scenarios. Besides, the novel approaches possess the talent to seek the solution to the inverse problem.}
\label{consGraphs4}
\end{figure*}

In Section~\ref{sect3}, sorting out the methodologies to solve PDEs with neural algorithms and their evolution has led to a certain understanding of the scenarios where different methods are applicable. 
Karniadakis et al.~\cite{KarniadakiGE2021PIML} have given a review of the Physics-Informed Machine Learning (PIML), including the spread applications after their proposal of PINNs. In this section, a comprehensive review of the various applications of both methodologies is provided (see Table~\ref{collectedpapers3} and Figure~\ref{consGraphs4}). 

\begin{table}[!htbp]
\begin{center}
\caption{Collected references for reviewing the actual applications of different approaches, scenarios addressed include the single-physics modeling of fluid, solid, heat, and electromagnetism etc., engineering and medical scenarios requiring multiphysics modeling, and inverse problem}
\label{collectedpapers3} \scalebox{1.0}{
\begin{tabular}{ccc}
\hline\hline
\rule{0pt}{12pt}
\multirow{2}{*}{\textbf{Applications}} & \multicolumn{2}{c}{\textbf{References}} \\ 
\cline{2-3}
\rule{0pt}{16pt}
& \textbf{PDEs as Constraints} & \textbf{Neural Operator} \\
\cline{1-3}
\rule{0pt}{16pt}
\textbf{Fluid} & \tabincell{c}{[2],[37],\\{[67]}\textasciitilde[80]} & [149]\textasciitilde[151] \\
\rule{0pt}{16pt}
\textbf{Solid} & [81],{[84]\textasciitilde[92]} & [152],[153]\\
\rule{0pt}{16pt}
\textbf{Heat} & [82],[93] & [158]\\
\rule{0pt}{16pt}
\textbf{Electromagnetism etc.} & [83],[94]\textasciitilde[98] & [154]\textasciitilde[157]\\
\rule{0pt}{20pt}
\tabincell{c}{\textbf{Engineering \&} \\ \textbf{Medical Scenarios}} & [99]\textasciitilde[131] & [158]\textasciitilde[169]\\
\rule{0pt}{18pt}
\textbf{Inverse Problem} & \tabincell{c}{[2],[37],[69],\\{[132]}\textasciitilde[148]} & [170]\textasciitilde[174]\\
\hline\hline
\end{tabular}}
\end{center}
\end{table}



\subsection{PDEs as Constraints for Single-Physics Modeling}
The development and utilization of CFD modeling are closely related to breakthroughs in science and engineering fields~\cite{BomphreyRJ2017SmartWing,BlockenB2007CFDSimulation}. Since this technique is still the guest of honor in high-end engineering~\cite{ChenL2012OFC}, simulations of hypersonic flow~\cite{Codoni2021Stabilizedmethods}, compressible flow~\cite{Yi2019Amulticomponent}, complex turbulent flow~\cite{Argyropoulos2015Recentadvances}, to name a few. Note that these techniques have been successfully utilized in modern aircraft. 
Also, CFD possesses one of the world's best-known and perhaps most complex modeling instruments, namely, the N-S equations~\cite{StokesGG1880firstNSEquations}. 
As a "reputable" system of PDEs, N-S equations can be employed to depict most of the scenarios in fluid mechanics. However, the application and solving conditions for practical problems are highly demanding. Simplifying or adding of more constraints must be first accomplished when wielding the equations in concrete terms~\cite{BhuttaMMA2012CFDapplications}. Thus, it is natural to consider whether a novel approach can be applied to solve this complex system. As of now, in addition to the extensive experiments that have been performed by Lagaris et al.~\cite{lagaris1998artificial} and Raissi et al.~\cite{Raissi2019Physics}, PINNs as well as other methods have been employed to study more challenging issues in fluid mechanics, including: 
\begin{itemize}
\item With the aid of physics-driven DNNs to build up the surrogate models as well as achieve the data assimilation for simple flows~\cite{SunLN2020Surrogatemodeling,BaiXD2020ApplyingPhysics,jin2021nsfnets}, such as the incompressible laminar flow~\cite{RaoCP2020PIDLforincompressible} that is often assumed in simulation scenes. The new method strikes a good balance between accuracy and computational cost;
\item Applying the approaches to deal with more complex scenarios, such as high-speed~\cite{mao2020physics} and turbulent fluid flows~\cite{WangR2020TowardsPIDL}, also includes specific cases like shock waves and bubble dynamics~\cite{WanZY2020PCML}. As of today, works against these issues remain the jewel in the simulation's crown. Despite the unique advantages of DL-based methods in solving inverse problems that will be spoken of later, their performance still needs to be improved in the face of these accuracy-sensitive computations;
\item Discovering the models of simple or turbulent flows from various scattered or noisy data with specially designed frameworks~\cite{RaissiM2019DLofTSM,MethaPP2019Discovering}. This class of utilization is also referred to as PIML~\cite{CruzMA2019UseofReynoldsForce,SunLN2020PcbayesianNN} and is more related to solving the inverse problem. In addition, novel approaches are adopted to perform the computing of multi-phase flow~\cite{Buhendwa2021InferringIncompressible}. In those ways, CFD problems can be studied at a deeper level or in a more engineering way~\cite{XuR2020WeakformTGNN}.
\end{itemize}


An evident question is: why are deep approaches still attempting to deal with CFD issues when traditional numerical methods have been already mature? The motivation is mainly two-fold. Firstly, from the perspective of building a surrogate  model~\cite{SunLN2020Surrogatemodeling}, taking PDEs as constraints is more efficient since they adopt DNNs to directly approximate the scenarios depicted by various constraints (PDEs and ICs/BCs), instead of repeatedly iterating at each time step to find the new equilibrium state as in traditional methods~\cite{WangR2020TowardsPIDL}; Secondly, ANNs, or more precisely, DNNs in their traditional applications tend to extract posterior information directly from the data to aid in inferring or predicting. A prominent barrier encountered in doing so is to search in vast spaces of potential parameters or architectures to achieve the target model~\cite{WanZY2020PCML}. By integrating the first principle thinking, i.e., introducing prior constraints inherent in the physical system and putting up an efficient framework with PI means, PIML incorporates the advantages of both direct approximation and posterior searching, and studies the scenes in forward and inverse perspectives~\cite{LuoSR2020ParameterIdentification}. 

In addition to utilization in CFD, some other single-physics modeling has employed new methods, such as solid mechanics~\cite{SamaniegoE2020AnEnergyApproach}, thermodynamics~\cite{ZobeiryN2021APIMLApproach}, and electromagnetic~\cite{BaldanM2021Solving1DNonlinear}. Tao et al.~\cite{TaoF2020PIANN}, Li et al.~\cite{LiW2021APGNNframework}, and Nguyen-Thanhang et al.~\cite{Nguyen-ThanhVM2020ADeepEnergyMethod} integrated deformation-related energy equations into loss functions of DNNs to enable efficient simulation of phenomena such as elastic deformation and bending of solids. Dehghani et al.~\cite{DehghaniH2021ANNaided}, Zhuang et al.~\cite{ZhuangXY2021DeepAutoencoder}, and Goswami et al.~\cite{GoswamiS2020AdaptiveFourthOrder} applied the approaches to poroelasticity, vibration analysis, and fracture modeling. Rad et al.~\cite{RadMT2020TheoryTraining} also tested the novel kits for issues related to metal solidification. 
Moreover, Physics- or Theory-Guided DNNs have also been adopted to characterize the metal bending and identify the fractures~\cite{LiuSM2021DLinSheetMetal,ShuklaK2020PINNforUltrasound}. This class of applications does not implant physical constraints as profoundly as other work but mainly integrates certain epiphenomenal laws.  
Since thermodynamic processes are often coupled with fluid or solid phenomena to solve engineering issues, it is not that meaningful to perform single-physics modeling~\cite{BorzdynskiOG2021ExploringDatasets}. 
Examples of Thermo-Hydro (TH) and more complex multiphysics simulations would arise later when it comes to more practical engineering scenarios. 
As for the modeling against electromagnetism~\cite{Noakoasteen2020PI}, optimal control~\cite{Nakamura-ZimmererT2021AdaptiveDLfor}, and even quantum mechanics~\cite{LiL2021KohnSham,BazarkhanovaA2021PINNsforSolvingDiracEquation}, several attempts have also been made up to now. 
It has been proven beneficial by combining mechanism and statistical perspectives for the microscopic physical world and complex systems~\cite{Razakh2021pnd}. 



\subsection{Engineering and Medical Scenarios}
As the frameworks to implement the prior knowledge in the form of PDEs have been well-studied in recent years, their potential applications in various engineering disciplines and fields have gradually emerged. In engineering scenarios dealing with the natural environment, such as the development of geological energy, Wang et al.~\cite{WangNZ2020DLofSubsurfaceFlow,WangNZ2021EfficientUncertainty} and Tartakovsky et al.~\cite{TartakovskyAM2020PIDNNsforLearningParameters} have focused on problem of the Hydro-Mechanical (HM) coupled simulation of subsurface flow. There are also several works attempt to achieve water resources observation~\cite{ChadalawadaJ2020HIML,BandaiT2021PINNsWithMonotonicity}. Since the internal structure of earth is highly complex and unknown to us, using the PIML method for data assimilation~\cite{WuHY2021PCDLforDataAssimilation,HeQZ2021PINNsforMultiphysics} can provide a good integration of engineering observations with known physical equations. In addition, the simulation of subsurface flow is also applicable to gases such as underground reservoir pressure~\cite{HarpDR2021OnTheFeasibility} and the response of the carbon dioxide storage site~\cite{ShokouhiP2021PIDL}. While at the surface, such simulation plays a vital role in wind energy generation~\cite{ZhangJC2021Spatiotemporal,ParkJY2019PIGNN} and climate prediction~\cite{SnaikiR2019KnowledgeEnhanced}. Similarly, novel kits have been hired in civil engineering~\cite{WangTS2021Reconstruction} and geohazard forecasting~\cite{KarimpouliS2020PIMLSeismic}, which deals with geotechnical and geological disciplines.

In high-tech and manufacturing industries, operation and production procedures in these fields have been artificially constrained and simplified compared to the natural world's complex and coupled physical processes, leading to more direct simulations. Yucesan et al.~\cite{YucesanYA2021Adjusting} and Wang et al.~\cite{WangJJ2020PGNN} were early to apply the method of PINNs to engineering mechanics and tool wear in manufacturing. Then, more attempts in segmentations were followed, such as the thermodynamics~\cite{MasiF2021Thermodynamics}/thermochemical curing process~\cite{NiakiSA2021PINNformodellingthethermochemical}, additive manufacturing\cite{ZhuQM2021MLforMAM}, and tool life~\cite{KarandikarJ2021PGLC}. By further expanding the idea, the above utilization also shows potential of the new methods in materials science~\cite{ZhangZZ2021PIDLforDigital,LiY2021DLbasedMethod}, acoustics~\cite{PettitCL2021APINNforSound}, and TH coupling~\cite{FlorioMD2021PINNsforRarefiedGas}. Meanwhile, the application of PI approaches to communication~\cite{HagerC2021PBDLforFiberOptic} and cybersecurity~\cite{NeuerMJ2020Quantifying}, and the Digital Twin~\cite{WangRH2020Kalibre} is of more interest.

As for the medical topics, which also include partial biological issues, modeling living organisms has been a long-standing wish in this field. Since an organism is so complex that some parts would be too tiny to acquire accurate results by purely instrumental measurements, it would be fascinate to build efficient models around this system while performing simulations and data assimilation based on them. Along these lines, Raissi et al. have applied PINNs to the microaneurysms simulation~\cite{raissi2020hidden,CaiSZ2021AIVelocimetry}. This computing of fluid dynamics for human tissue is also considered one of the most representative applications to date.
Simultaneously, modelings on thrombus~\cite{YinML2021NonInvasive} and cardiovascular~\cite{KissasG2020MLinCardiovascular,FossanFE2021MLAugmented}, as well as more studies on biological soft tissue~\cite{LiuML2020AGenericPINN} and wound healing~\cite{LagergrenJH2020BiologicallyInformed}, have shown that DL based on PDEs are becoming powerful assistance in medical and biological scenarios. Overall, promoting the fusion and practice of new approaches with multiphysics simulation would be necessary to boost the development of solving PDEs with DNNs. More importantly, relying on them to achieve multiphysics modelings in engineering and medical applications is also considered the way to go.



\subsection{Inverse Problems}
A different reason to consider the simulation of microaneurysms representative is that this move also integrates another vital form of utilizing PDEs: solving inverse problems. As quoted in the work of Jin et al., the need for solving inverse problems is widespread in scientific and engineering research\cite{jin2021nsfnets}, and it is mainly due to two reasons:
\begin{itemize}
\item Limited by the current development of measurement technology, there are some physical phenomena/quantities that the installed instruments cannot directly observe. Scientists and engineers often have to derive the values of the target physical quantity by measuring the indirect one\cite{ZhangYJ2021HiddenPhysics}. A typical example is inversions emphasized in geophysical research;
\item Ideal or simplified mathematical and physical modeling can still deviate from the evolution of the real world, e.g., there are generally huge gaps between models of different scales and fidelity\cite{ChakrabortyS2021TLbasedMFPIDNN}. Even though researchers can now build up a complex set of PDEs and solve them jointly, some of the information missing in mechanistic modeling is still not covered.
\end{itemize}

As a result, how to invert the physical quantities from obtained observations or calibrate the parameters based on data to get models more consistent with reality becomes the most concern in studying inverse problems.
Previous studies relying mainly on numerical methods, the solution to the inverse problem was very challenging since it commonly requires a tedious framework to accomplish the targeted study~\cite{Isakov2006InverseProblems}. However, it becomes less complicated with the NN-based approaches. In the most mainstream applications, the fundamental task of DL represented by NNs is to establish an excellent mapping between the input and output data and thus extract the hidden patterns in the data. Currently, novel approaches incorporating AI and SC become a better choice after training a DNN as the pre-trained model with prior physical knowledge and then putting it into the original advantageous area. Relying on massive data to achieve the fine-tuning makes the model more precise and realistic. 


Attempts were made in various aspects of scientific research and engineering practices, including Chen et al.'s work on nano-optics and metamaterials\cite{ChenYY2020PINNforInverseProblems} and Lu et al.'s\cite{LuL2020Extraction} work on extracting materials' mechanical properties, as well as Bekele's\cite{BekeleYW2020PIDL} improving 1-D consolidation model. In addition, there are also trials made by Lai et al.\cite{LaiZL2021StructuralIdentification} on structural identification. In the typical scenarios of inverse problems, i.e., previously mentioned geophysics and CFD modeling, PI methods are hired to detect underwater obstacles\cite{KahanaA2020ObstacleSegmentation} and reconstruct subsurface flow\cite{WangNZ2021DLBIMA}, complex fluid\cite{MahmoudabadbozchelouM2021DataDriven}, and HM coupling\cite{RaissiM2019DLofVortexInducedVibrations}. Similarly, novel approaches have been used in reconstructing the ultrasound elasticity images\cite{MohammadiN2021Ultrasound}. 
As an emerging discipline that has only been developed for more than a hundred years, many models in the field of quantum physics are subject to the second issue mentioned earlier\cite{MishraS2021PINNsforSRT}. Silva et al. and Zhou et al. hence introduced the Physics-Guided methods to correct the Eikonal Equation\cite{SilvaRM2020PINNsfortheFEE} and Schrödinger equation\cite{ZhouZJ2021SolvingForwardandInverseProblems}. At the same time, Parkravan et al.\cite{PakravanS2021SolvingInversePDE} and Meng et al.\cite{MengXH2021MultiFidelity} not only applied the PINNs method to solve the inverse problem but also improved it with the Bayesian strategy.  


\subsection{Utilization of NOs}
In contrast to the approaches treating PDEs as training constraints, utilization of the NO methodology is still in its infancy. Although the mathematical prototype of DeepONet was proposed and systematically studied by Chen et al. as early as 1993~\cite{chen1993approximations}, improvement of the computational performance and the implementing techniques have not caught up with the theories until recent years. Then, the NO methods revealed their power only after two papers on DeepONet\cite{Lu2021Learning} and FNO\cite{LiZY2021Fourier} were published in early 2021.
Especially from late 2021 to early 2022, novel approaches based on functional approximation were practiced in several fields, such as classical CFD issues\cite{UsmanA2021MLCFD} like bubble growth\cite{LinCS2021Aseamlessmultiscale},\cite{LinCS2021Operatorlearning} and mechanical problems\cite{YinML2022Interfacing} like elastic waves\cite{Zhang2022Tianze}. Topics related to electromagnetism\cite{CaiSZ2021DeepMMnet},\cite{ZhuCK2021FastSolver} and dynamical systems\cite{WangYF2021Modelingofnonlineardynamic},\cite{YinY2021LEADS} are not left out. Meanwhile, engineering applications are also in high demand. Different NO methods were utilized in predicting the performance of solar-thermal systems\cite{OsorioJD2022Forecasting}, the simulation of 
carbon dioxide multi-phase flows\cite{YanBC2022Arobustdeeplearning}, the study of heterogeneous porous materials\cite{NASEM2021MLforHetergeneous} in the energy field, and geophysical\cite{WangLM2022Progressandprospect} and climate forecasting\cite{PetersenK2021AdvancingSeaIce},\cite{LeungLYR2021PIL},\cite{PeregoM2021Hybrid}.  
Since it possesses the capability to incorporate multi-precision and multi-scale data into the modeling process, the new approach is certainly more efficient and promising than just adopting a fixed set of PDEs (physical laws and ICs/BCs) as constraints.
In addition, in communication techniques, financial engineering, and cutting-edge research, NO methods have been applied to identify specific emitters\cite{ZhaX2022SpecificEmitter}, correct financial models\cite{LeiteIMS2021TheDeepONetsforFinance}, simulate supersonic issues\cite{MaoZP2021DeepMMnet} and turbulent combustion\cite{GitushiKM2022Investigation}, and support aerospace activities\cite{ShiRH2021Metamodel}.  Last but not least, their solution of the inverse problem\cite{MacKinlayD2021ModelInversion} is making a splash in geophysics\cite{YangY2021SeismicWave},\cite{SongC2022Highfrequency} and water resources observation\cite{BrelsfordC2021AIImproved},\cite{OzbayAG2022DLfluidflow}.

\section{Discussion}\label{sect5}
Computer-implemented solution methods possess a common feature: the target solutions would be approximated by computational solutions rather than obtaining the analytical solutions directly~\cite{heath2018scientific}. Traditional numerical means such as FDM/FEM and the novel approaches of taking PDEs as constraints and NOs all embody this thought. However, applying DNNs to solve PDEs is undoubtedly more efficient and thorough: the well-designed networks are trained to approximate the target solutions in multiple dimensions simultaneously, even in the time dimension. Unlike traditional numerical methods, DNN-based methods can learn the direct mappings between physical state and spatial/temporal coordinate without repeatedly iterating at each time step.

\begin{figure}[htbp]
\centering
\includegraphics[width=0.48\textwidth]{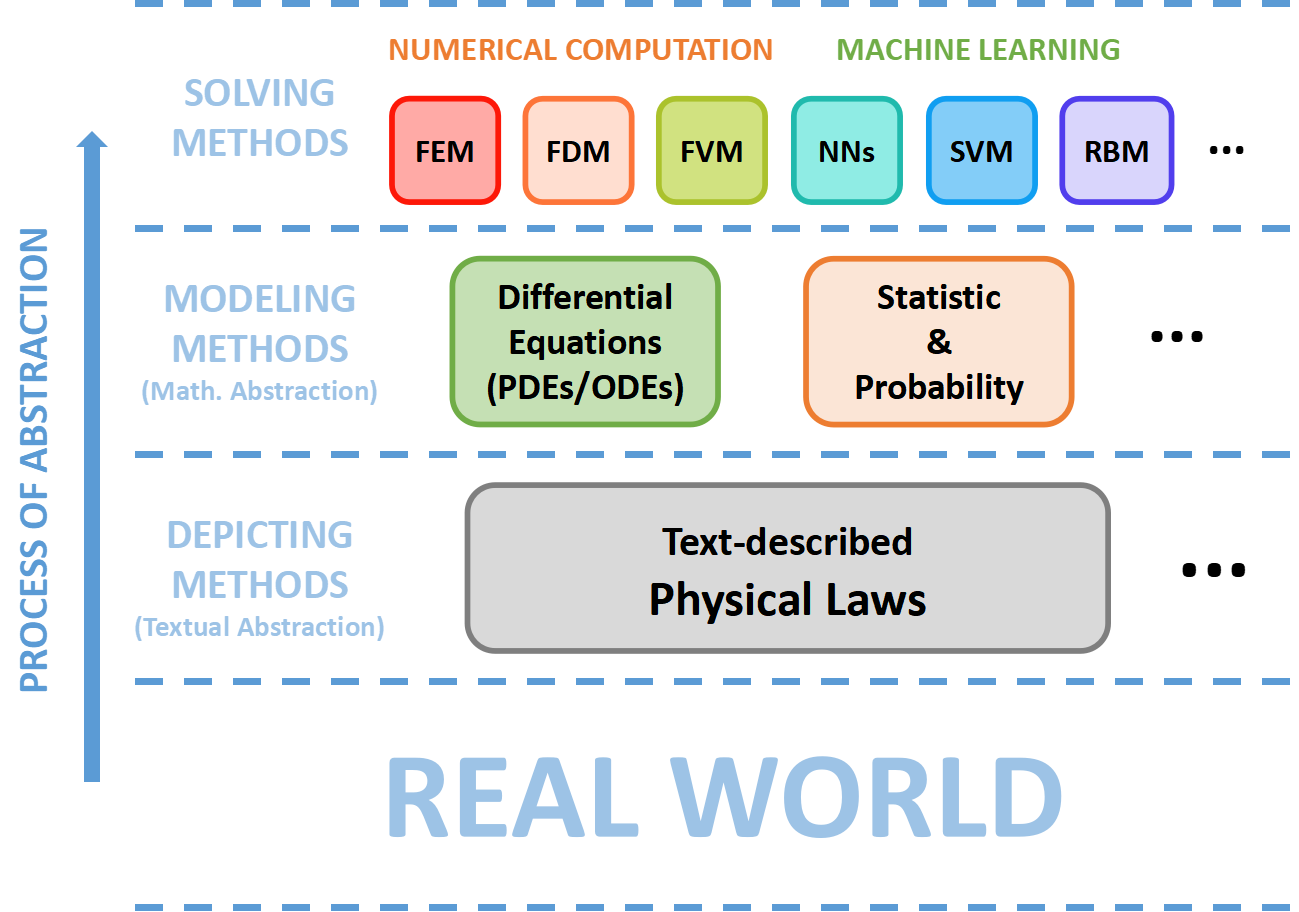}
\caption{Levels and categories of the different approaches for depicting, modeling, and solving the scientific as well as engineering issues in the real world.}
\label{consGraphs6}
\end{figure}

\subsection{Methodology Portfolio}
Further comparing the novel methodologies with numerical ones, the instrument used to depict the scenarios has not changed and remains the DEs. At the same time, appliances for solving the model have been updated to DNNs. Hence, a comprehensive review of the depicting, modeling, and solution methods would certainly be essential when it comes to the utilization and improvement of their portfolio. 
According to Dr. Jim Gary, scientific research can be classified into four paradigms: experiment, theory, simulation, and data science~\cite{HeyT2009TheFourthParadigm}, as shown in Figure~\ref{consGraphs6}. 
From the abstraction of real-world to the solution of models can be roughly put into four levels~\cite{farlow2006introduction}: 
\begin{itemize}
\item The first level is the real world itself, which is complex and rich in variation;
\item In the second level, after summing up rules in the real world based on experiments or experience, the physical laws were tried to be expressed in words, e.g., a literal version of Newton's law: \textit{a body remains at rest, or in motion at a constant speed in a straight line, unless acted upon by a force}. This level also encompasses the experimentation paradigm and a part of the  theory paradigm;
\item Mathematics instruments were gradually developed as human society progressed. In the third level, describing those laws by means of mathematical methods such as DEs and probability makes the formulation more efficient and precise. This evolution has boosted the last formation of the theoretical paradigm;
\item Finally, applying the laws to predict unknown phenomena has greater significance than merely depicting the rules. However, the solution of the descriptive model is essential to accomplish this process. And in this level, the application of different approaches has bifurcated.
\end{itemize}

Up to now, the solution methods could be broadly classified into numerical computation, machine learning, etc., according to their characteristics. The numerical analysis mainly aims to directly depict the mechanism with DEs, while other methods represented by ML are primarily oriented toward probabilistic statistics that characterize the data. In the four-paradigm system, they also represent the last two paradigms, namely simulation and data science. 
However, when handling a specific problem with solution methods, it can actually be changed or even interchanged if the conditions are suitable since every means essentially tries to approximate the goals or the reality. For example, as previously elaborated, the training of DNNs can be constrained by ODEs/PDEs to match the scenarios described by them, i.e., II in Figure~\ref{consGraphs7}. Further, specially designed deep networks, such as DeepONet and FNO, can also be employed to fit the functional describing the physical mechanisms contained in a large amount of data (I in Figure~\ref{consGraphs7}), which is certainly advantageous in solving multi-scale and multi-fidelity issues. On the contrary, one can also reverse the operation, e.g., applying the patterns of the dynamical system to enhance the DNNs (III in Figure~\ref{consGraphs7})~\cite{HuPP2022RevealingHidden}. In that case, it dramatically benefits improving the interpretability of Deep Learning. 

\subsection{New Perspectives}
As mentioned before, in addition to the initial derivation of the relevant mathematical theories, scientists have also devoted themselves to analyzing the generalization errors of methods such as PINNs\cite{MishraS2022EstimatesOnTheGeneralization},\cite{mishra2022estimates} and DeepONet\cite{Kovachki2021OnUniversal},\cite{Lanthaler2022Errorestimates}. These methods have proven to have excellent performance for the work they are going to accomplish. Further, Lu et al. have also compared the performance of two different approaches~\cite{LuL2022Acomprehensive}. More often, there is a desire to further improve the existing methods based on former experience, e.g., PINNs are at the forefront. As shown in Figure~\ref{consGraphs5} and~\ref{consGraphs7}, an overall architecture of training NNs to solve PDEs can actually be viewed as consisting 
of the input, NNs for approximation, and the constraints imposed on NNs through the loss function. 
Then, one can start from any of the three aspects when it comes to improving performance: 
\begin{figure}[htbp]
\centering
\includegraphics[width=0.48\textwidth]{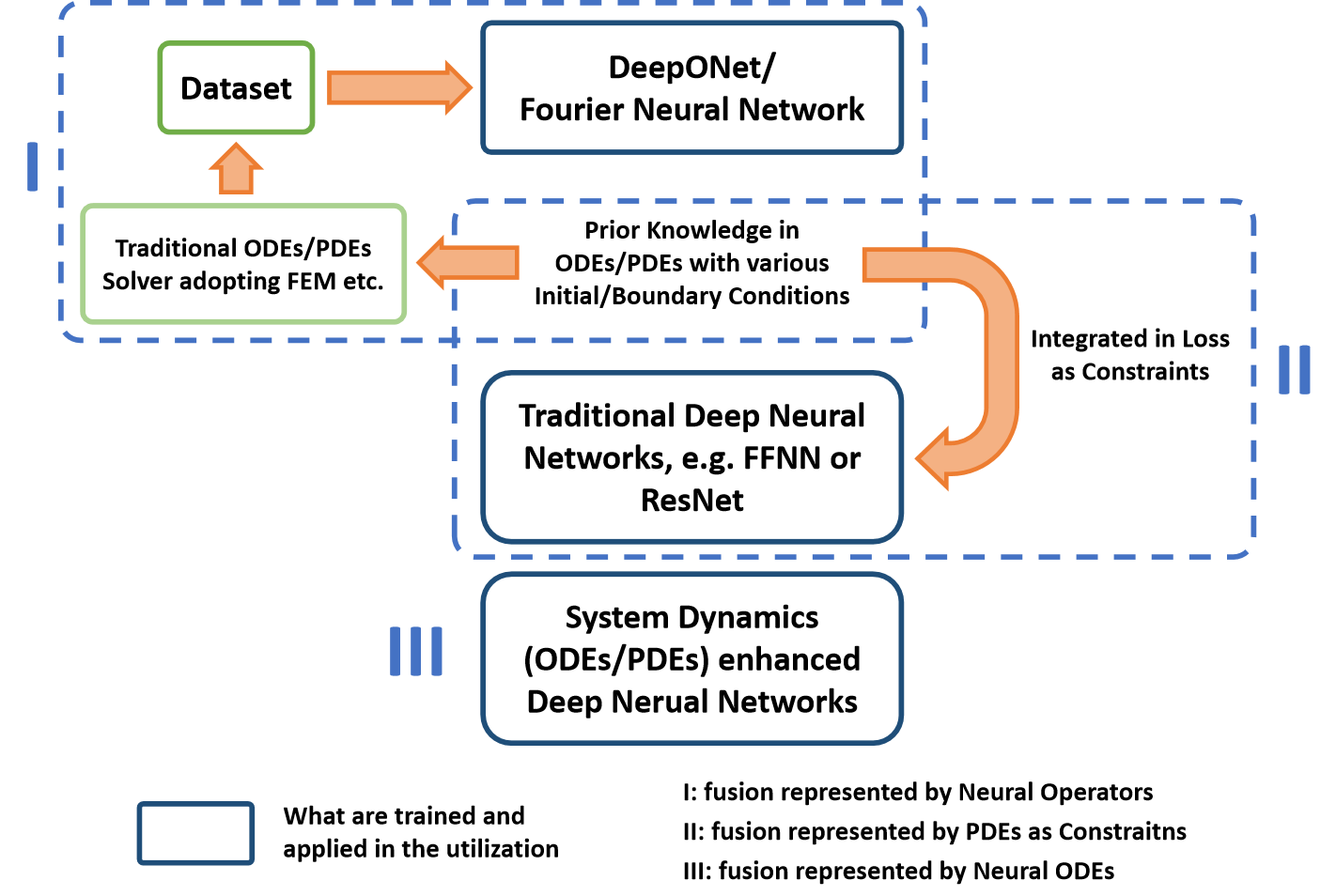}
\caption{Three forms of fusion between Deep Learning and PDE-based SC: I. applying various specially designed deep networks, e.g., DeepONet and FNO, to fit the functionals that can depict different physical issues or practical scenarios; II. taking DEs as constraints to train DNNs, thus obtaining the surrogate models; III. employing the known patterns of the dynamical system to enhance the performance of DNNs.}
\label{consGraphs7}
\end{figure}
\begin{itemize}
\item {\bf{Loss function:}}
reforming the loss functions is often considered to enable the acceleration of training works. Jagtap et al. further developed the conservative and extended PINNs (cPINNs~\cite{jagtap2020conservative} and XPINNs~\cite{JagtapAD2020XPINNs}) by constructing multiple sub-domains and adding constraints to the flux conductivity on the interfaces. Meanwhile, Kutz et al. and Patel et al. have achieved the same goal by introducing parsimony~\cite{KutzJN2022Parsimony} and spatio-temporal control volume scheme~\cite{PatelRG2022Thermodynamically} as regularizers. Besides, several studies have also worked on utilizing the given initial and boundary conditions in a different way, e.g., Li et al.~\cite{LiC2022LearningHighOrder} have combined the PI approach with the level set method;
\item {\bf{DNNs:}} 
involving various types of DNNs can also benefit from their inherent areas of strength. That is, beside FFNNs and the bifurcated structure, different DNN frameworks can be adopted~\cite{JagtapAD2022DeepKNNs,SirignanoJ2018DGM,GuptaG2021Multiwavelet}. For example, RNNs and Generative Adversarial Networks (GANs), have improved the capability of both methodology to learn features in the advantageous domains of the network, i.e., the sensitivity of RNN/LSTM-NNs/Transformer~\cite{RenP2022PhyCRNet,delAguilaFerrandisJ2021LearningFunctionals,CaoSH2021Choose} to temporal data, and the talent enhanced by GANs~\cite{GaoYH2022Wasserstein} to solve the Stochastic Differential Equation (SDE)~\cite{MengXH2022LearningFunctionalPriors}. U-FNO with additional U-Fourier layers~\cite{WenG2022UFNO} and Attention-enhanced FNO~\cite{PengWH2022AttentionEnhanced} were also proposed, with reformed FNOs gaining improved training accuracy and more capable of complex scenarios such as multi-phase flow. Moreover, attempts to optimize the activation functions~\cite{WangSF2021OnTheEigenvector} and build evolvable deep networks are also carefully designed. Parameters of the evolutional DNNs~\cite{DuYF2021EvolutionalDNN} can be dynamically updated during utilization;


\item {\bf{Input:}} since the basic idea of Physics-Guided methods is to constrain the training through loss functions, an improvement from the input side is thought of as not being that cost-effective. However, PINNs' retaining the ease of data-driven approximation makes this idea feasible, e.g., Kovacs et al. included parameters that determine the coefficients of the eigenvalue equation as an additioninput~\cite{KovacsA2022ConditionalPINNs}. Moreover, the same idea would be more favorable for enhancing NOs, e.g., Hadorn has improved the structure of DeepONet to allow the basis functions to be shifted and scaled by the values obtained from the input functions~\cite{Hadorn2022ShiftDeepONet}. Lye et al.~\cite{LyeKO2021AMultiLevelProcedure} proposed to improve the accuracy of ML algorithm by multi-level procedure;
\end{itemize} 

Another area where DL is recognized as superior for solving DEs is that its properties are well suited for solving high-dimensional PDEs, thus could break the notorious "curse of dimensionality"~\cite{han2018solving}. Hence, in the last year or two, there have been attempts to seek improvement in a compound way. Hutter et al.~\cite{HutterC2022MetricEntropy} employed RNNs to accomplish the approximation of general discrete-time linear dynamical systems. Zeng et al.~\cite{ZengSJ2022AdaptiveDNNs} proposed to enhance the training process of DNNs from three perspectives: adaptive choosing the loss function, adaptive selecting the activation function, and adaptive sampling. Their work has achieved promising results, especially for high-dimensional PDEs. Beyond that, adapting the model's training process based on experience from DL is another trial direction, e.g., Wang et al.~\cite{WangSF2022WhenAndWhy},~\cite{WangSF2021UnderstandingAndMitigating} and Mattey et al.~\cite{MatteyR2022ANovelSequentialMethod} have conducted a number of studies in this field. 

However, a more meaningful way would be the combination of both methodology and other methods. For example, Wang et al.~\cite{WangSF2021LearningTheSolution} and Patel et al.~\cite{PatelRG2021APIOperatorRegression} proposed to use PDEs as constraints for training DeepONet. This means that instead of the posterior information brought by inputs, Physics-Informed DeepONet could learn the prior knowledge directly from PDEs~\cite{GoswamiS2022APIVariationalDeepONet}. This aggregation improves the training efficiency while avoiding relying on the traditional simulator that generates the training data. The same idea can certainly be employed for the reforming of FNO. 
In addition, various approaches combining Extreme Theory of Functional Connections (X-TFC) or ConvPDE-UQ with PINNs and DeepONet have been proposed by Schiassi et al.~\cite{SchiassiE2021ExtremeTheory} and Winovich et al.~\cite{WinovichN2021NNApproximation} with positive performance.

As a field with great potential for development and application, works applying DL to the solution of PDEs have blossomed in recent years. We believe there are some directions that worth to be focused on in the future. 
First, regarding high-end simulation, the main issue currently hindering the further application of PI methods remains its weak point of accuracy compared to numerical methods. However, practical computing in engineering scenarios is commonly resource-constrained and has less extreme claims to accuracy compared to the scientific research supported by supercomputers. Thus, engineering application with targeted specialization is considered a good entry point for new approaches. Moreover,  
an approximation to the functionals makes the NO method much less constrained in updating the modeling scenes than the former. It would hence be a good idea to differentiate the development of the two: the Physics-Guided approaches focus on the solution of high-dimensional PDEs, while the NOs orient to the migratable modeling, e.g., building up libraries of pre-trained models and then fine-tuning them for concrete scenarios. Finally, early development in the fusion of DL and SC is mainly driven by physical and mathematical research with relevant demands. From our side, the potential of ANNs and DL themselves remains to be explored. New methodologies can be introduced into other parallel topics besides relying on a thorough understanding of ANNs to improve computing accuracy further, e.g., 
the formularized prior knowledge in PDEs can be beneficial in dealing with scenario-specific multimodal coordinated representations, especially in the scenes of scientific experiments with certain theoretical foundations to help improve the physical/mathematical models.




\section{Conclusion}\label{sect6}
In this paper, we have reviewed and carefully sorted out the rise of a subdivision of AI4Science, namely the use of DNNs to solve PDE-related issues. It includes the proposal and theoretical development of approaches to solving PDEs with DNNs for decades, their gradual implementation, and the blossoming of applications in recent years. It can be found that the fusion of SC with DL is inevitable because of the advantages brought by theoretical and technical development. On the other hand, due to the great abundance of computational capability catalyzing an urgent need for new research paradigms, this trend is reaching its flash point. 
Specifically, the last gaps between mechanism models based on experiments or expert experience and the real world are as ghostly dark matter and difficult to eliminate through available means. Similarly, although data-driven DL exhibits powerful pattern extraction capabilities, interpretability issues follow up. 
New achievements, represented by PI and NO methods, bridged those gaps and weak points by integrating the prior knowledge from PDEs and the posterior information from massive data into DL.

The robust developing instruments available today also assiduously transform them into research and production. 
As mentioned before, Lu et al. developed DeepXDE~\cite{Lu2021DeepXDE} involving PINNs and DeepONet. In industry, NVIDIA, known internationally for GPU and cuda toolkit, has integrated PINNs, DeepONet, and enhanced FNO in its Modulus framework (formerly SimNet) and applied this powerful toolbox to the building of the Digital Twin. Even the latest quantum information science has cast a favorable eye on NO~\cite{SwanM2021QuantumInformationScience}. Besides, more researchers are entering this field with more innovative ideas~\cite{WangHJ2022MosaicFlow}, e.g., introducing Gaussian processes~\cite{ChenYF2021SolvingAndLearningNonlinearPDEs} and constructing hybrid FEM-NN models~\cite{MituschSK2021HybridFEMNN}.
Recently, Li et al. built on their previous works of FNO and proposed a DL framework specifically for the learning of operators~\cite{DBLP:journals/corr/abs-2108-08481}. One of the highlights of the newly developed framework is the proven property of discretization-invariance, undoubtedly having great significance for its wider application.
Also, as stated in the earliest work of DeepONet~\cite{Lu2021Learning}, there has been an outlook on introducing the latest DL methods and techniques into this burgeoning field.
Although these new approaches are various, they all have one thing in common. Namely, the boundaries between mechanism and data are no longer stuck to, but the advantages of both sides are organically merged to achieve a breakthrough. Last but not least, what is more anticipated, of course, is a more profound fusion~\cite{NelsenNH2021TheRandomFeature},~\cite{KadeethumT2021AFrameworkForDataDriven},~\cite{GuptaR2022AHybridPartitioned},~\cite{GinCR2021DeepGreen},~\cite{BaoG2020NumericalSolution},~\cite{Jin2022MIONet},~\cite{Lu2022MultifidelityDNO}.

\section*{Acknowledgments}
We thank the anonymous associate editor and reviewers for their helpful comments and suggestions.

\bibliographystyle{IEEEtran}
\bibliography{3_reference}

\end{document}